\documentclass[conference]{IEEEtran}
\usepackage{times}
\usepackage{overpic} 
\usepackage{graphicx} 
\usepackage[ruled,linesnumbered]{algorithm2e}
\usepackage{amsmath}
\usepackage{amssymb}
\usepackage{multirow}
\usepackage{booktabs}
\usepackage{xspace}
\usepackage[noend]{algpseudocode}
\usepackage{makecell}
\usepackage{cite}  
\UseRawInputEncoding
\usepackage[utf8]{inputenc}
\usepackage[table]{xcolor}
\usepackage{colortbl}
\usepackage{array}
\usepackage{hyperref}
\usepackage[table]{xcolor}   
\usepackage{pgf}               
\usepackage{etoolbox}          
\usepackage[table]{xcolor}
\usepackage{booktabs}
\usepackage{multirow}
\usepackage{pifont}
\usepackage{array}
\usepackage{color, colortbl} 
\usepackage{booktabs}   
\usepackage{multirow}   
\usepackage{graphicx}   
\usepackage{xcolor}     
\usepackage{colortbl}   
\usepackage{pifont}     
\usepackage{arydshln}   
\usepackage{subcaption}
\newcommand{\cmark}{\ding{51}}
\newcommand{\xmark}{\ding{55}}
\definecolor{green1}{RGB}{120, 198, 121} 
\definecolor{green2}{RGB}{173, 221, 142} 
\definecolor{green3}{RGB}{217, 240, 199} 
\definecolor{yellow1}{RGB}{255, 237, 160} 
\definecolor{yellow2}{RGB}{254, 217, 118} 
\definecolor{orange1}{RGB}{253, 174, 107} 
\definecolor{orange2}{RGB}{253, 141, 90}  
\definecolor{red1}{RGB}{252, 108, 96}     
\definecolor{red2}{RGB}{227, 74, 74}      

\newcommand{\getcolor}[1]{%
    \ifdim #1pt < 0.03pt \cellcolor{green1}%
    \else\ifdim #1pt < 0.05pt \cellcolor{green2}%
    \else\ifdim #1pt < 0.07pt \cellcolor{green3}%
    \else\ifdim #1pt < 0.09pt \cellcolor{yellow1}%
    \else\ifdim #1pt < 0.12pt \cellcolor{yellow2}%
    \else\ifdim #1pt < 0.15pt \cellcolor{orange1}%
    \else\ifdim #1pt < 0.20pt \cellcolor{orange2}%
    \else\ifdim #1pt < 0.30pt \cellcolor{red1}%
    \else \cellcolor{red2}%
    \fi\fi\fi\fi\fi\fi\fi\fi%
}

\newcommand{\getcol}[3]{%
    \pgfmathsetmacro{\normalized}{(#1-#2)/(#3-#2)}%
    \ifdim \normalized pt < 0.11pt \cellcolor{green1}%
    \else\ifdim \normalized pt < 0.22pt \cellcolor{green2}%
    \else\ifdim \normalized pt < 0.33pt \cellcolor{green3}%
    \else\ifdim \normalized pt < 0.44pt \cellcolor{yellow1}%
    \else\ifdim \normalized pt < 0.55pt \cellcolor{yellow2}%
    \else\ifdim \normalized pt < 0.66pt \cellcolor{orange1}%
    \else\ifdim \normalized pt < 0.77pt \cellcolor{orange2}%
    \else\ifdim \normalized pt < 0.88pt \cellcolor{red1}%
    \else \cellcolor{red2}\fi\fi\fi\fi\fi\fi\fi\fi%
}

\newcommand{\getsem}[3]{%
    \pgfmathsetmacro{\normalized}{(#1-#2)/(#3-#2)}%
    \ifdim \normalized pt > 0.89pt \cellcolor{green1}%
    \else\ifdim \normalized pt > 0.78pt \cellcolor{green2}%
    \else\ifdim \normalized pt > 0.67pt \cellcolor{green3}%
    \else\ifdim \normalized pt > 0.56pt \cellcolor{yellow1}%
    \else\ifdim \normalized pt > 0.45pt \cellcolor{yellow2}%
    \else\ifdim \normalized pt > 0.34pt \cellcolor{orange1}%
    \else\ifdim \normalized pt > 0.23pt \cellcolor{orange2}%
    \else\ifdim \normalized pt > 0.12pt \cellcolor{red1}%
    \else \cellcolor{red2}\fi\fi\fi\fi\fi\fi\fi\fi%
}

\newcommand{\applycolor}[1]{%
    \ifdim #1pt < 10.0pt \cellcolor{red2}
    \else\ifdim #1pt < 15.0pt \cellcolor{red1}
    \else\ifdim #1pt < 20.0pt \cellcolor{orange2}
    \else\ifdim #1pt < 25.0pt \cellcolor{orange1}
    \else\ifdim #1pt < 30.0pt \cellcolor{yellow2}
    \else\ifdim #1pt < 40.0pt \cellcolor{yellow1}
    \else\ifdim #1pt < 50.0pt \cellcolor{green3}
    \else\ifdim #1pt < 60.0pt \cellcolor{green2}
    \else \cellcolor{green1}
    \fi\fi\fi\fi\fi\fi\fi\fi%
}
\newcommand{\applycolorfloat}[1]{%
    \ifdim #1pt < 0.05pt \cellcolor{red2}
    \else\ifdim #1pt < 0.10pt \cellcolor{red1}
    \else\ifdim #1pt < 0.15pt \cellcolor{orange2}
    \else\ifdim #1pt < 0.20pt \cellcolor{orange1}
    \else\ifdim #1pt < 0.25pt \cellcolor{yellow2}
    \else\ifdim #1pt < 0.30pt \cellcolor{yellow1}
    \else\ifdim #1pt < 0.50pt \cellcolor{green3}
    \else\ifdim #1pt < 0.70pt \cellcolor{green2}
    \else \cellcolor{green1}
    \fi\fi\fi\fi\fi\fi\fi\fi%
}

\definecolor{bestgreen}{RGB}{198, 234, 198} 
\definecolor{goodyellow}{RGB}{255, 248, 190} 
\definecolor{slamblue}{RGB}{200, 230, 230} 

\begin{document}

\title{IRIS-SLAM: Unified Geo-Instance Representations for Robust Semantic Localization and Mapping}


    
\author{
    \authorblockN{
        Tingyang Xiao$^{1}$, 
        Liu Liu$^{1}$, 
        Wei Feng$^{1}$, 
        Zhengyu Zou$^{1}$, 
        Xiaolin Zhou$^{1}$, \\
        Wei Sui$^{2}$, 
        Hao Li$^{3}$, 
        Dingwen Zhang$^{4}$, 
        and Zhizhong Su$^{1}$
    }
    \authorblockA{
        $^{1}$Horizon Robotics \quad
        $^{2}$D-Robotics \\
        $^{3}$S-Lab, Nanyang Technological University \\
        $^{4}$Hefei Comprehensive National Science Center, Institute of Artificial Intelligence 
    }
}


%

\maketitle

\begin{abstract}
Geometry foundation models have significantly advanced dense geometric SLAM, yet existing systems often lack deep semantic understanding and robust loop closure capabilities. Meanwhile, contemporary semantic mapping approaches are frequently hindered by decoupled architectures and fragile data association. We propose IRIS-SLAM, a novel RGB semantic SLAM system that leverages unified geometric-instance representations derived from an instance-extended foundation model. By extending a geometry foundation model to concurrently predict dense geometry and cross-view consistent instance embeddings, we enable a semantic-synergized association mechanism and instance-guided loop closure detection. Our approach effectively utilizes viewpoint-agnostic semantic anchors to bridge the gap between geometric reconstruction and open-vocabulary mapping. Experimental results demonstrate that IRIS-SLAM significantly outperforms state-of-the-art methods, particularly in map consistency and wide-baseline loop closure reliability.
\end{abstract}

\IEEEpeerreviewmaketitle

\section{Introduction}

Simultaneous Localization and Mapping (SLAM) is a fundamental capability for autonomous navigation and high-level task execution. Beyond accurate pose estimation, practical SLAM systems must support reliable re-localization and object-level reasoning in complex environments. In these scenarios, purely geometric reconstruction is often insufficient, as it fails to capture the semantic context necessary for deep scene understanding.

Recent feed-forward 3D reconstruction models, such as DUSt3R \cite{dust3r_cvpr24}, MASt3R \cite{mast3r_eccv24}, VGGT \cite{wang2025vggt}, and DepthAnything3 \cite{depthanything3}, have revolutionized dense scene reconstruction. Unlike methods relying on local feature correspondences, these models learn a shared latent representation that enforces geometric consistency across views, enabling stable estimation even under severe viewpoint changes. While recent systems like MASt3R-SLAM \cite{murai2024_mast3rslam}, VGGT-SLAM \cite{maggio2025vggt-slam}, and VGGT-Long \cite{deng2025vggtlongchunkitloop} successfully leverage these advances, they remain predominantly geometry-centric. Crucially, visual observation is inherently holistic, encoding both geometric structure and rich semantic context. While feed-forward models implicitly capture this unified information within their latent representations, current pipelines effectively discard the semantic half. They restrict the learned cross-view consistency to geometric tasks, leaving the rich semantic cues unexploited for state estimation.

\begin{figure}[h]
\centering
\vspace{-4pt}
\includegraphics[width=1.0\linewidth]{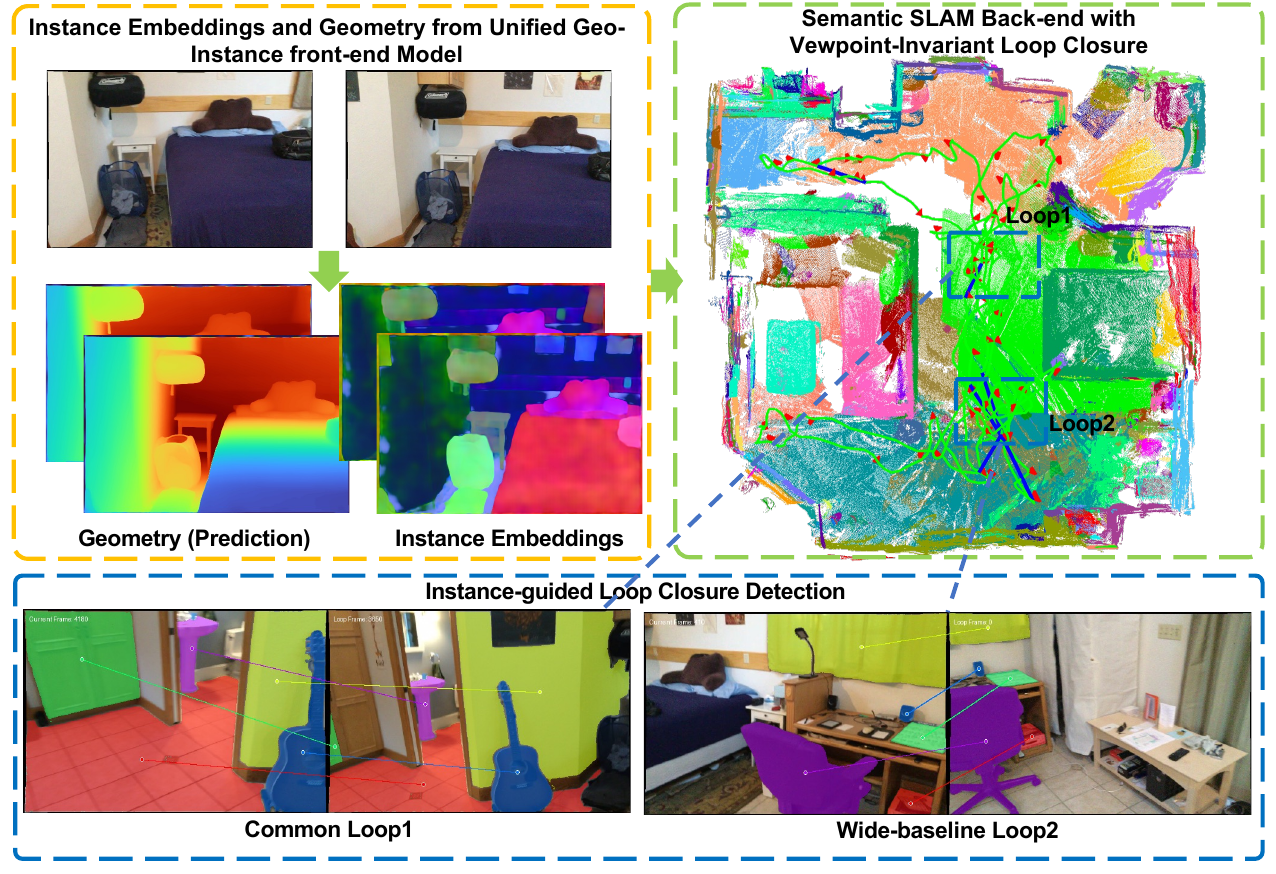}
\caption{Pipeline of IRIS-SLAM. The system uses a \textbf{Unified Geo-Instance Front-end} to jointly infer high-fidelity depth and cross-view consistent instance embeddings from monocular RGB streams (\textbf{Left}). These outputs are synergized to construct globally consistent dense semantic maps (\textbf{Right}). The viewpoint-agnostic instance embeddings act as stable semantic anchors, enabling robust loop closure detection under extreme wide-baselines and challenging perspective changes (\textbf{Bottom}).}
\vspace{-8pt}
\label{fig:teaser}
\end{figure}

In parallel, open-vocabulary semantic mapping approaches have enabled zero-shot 3D perception by combining 3D reconstruction with vision-language models such as CLIP \cite{radford2021learningtransferablevisualmodels}. These methods construct semantic maps either from reconstructed point clouds \cite{Peng2023OpenScene,openmask3d,nguyen2023open3dis} or from RGB-D observations \cite{conceptfusion,gu2024conceptgraphs,jin2025openfusionopenvocabularyrealtimescene,xia2024opengraphopengraphfoundation}. While recent SOTA frameworks like OmniMap \cite{OmniMap} and RAZER \cite{patel2025razerrobustacceleratedzeroshot} have pushed the boundaries of multimodal fusion and spatio-temporal aggregation, they still exhibit two fundamental limitations. First, \textbf{architectural decoupling}: semantic mapping is typically performed as an offline post-processing step dependent on pre-computed poses. This prevents semantic information from correcting the SLAM front-end in real time. Second, \textbf{fragile data association}: instance association relies primarily on geometric overlap or heuristic thresholds. This dependence makes the semantic maps sensitive to geometry errors, leading to fragmented representations when the underlying geometry degrades. Consequently, the semantic layer remains a passive artifact rather than a reliable cue for robust estimation.

For the same underlying reason, visual loop closure detection remains a persistent challenge. Traditional appearance-based approaches, such as ORB with Bag-of-Words (BoW) \cite{ORBfeature, Bow}, and global descriptors including NetVLAD \cite{arandjelović2016netvladcnnarchitectureweakly} or SALAD \cite{Izquierdo_CVPR_2024_SALAD}, rely on holistic image similarity. Under extreme viewpoint changes or wide-baseline revisits, pixel-level appearance similarity no longer reflects the shared scene structure, causing these methods to fail. This limitation again highlights the absence of stable, view-invariant entities that can anchor cross-view association over time.

We observe that the limitations in semantic mapping and loop closure share a common root cause: the lack of persistent, cross-view consistent entities in the estimation pipeline. Both tasks fundamentally require instance-level anchors that remain stable across viewpoint, illumination, and partial observations. Crucially, feed-forward reconstruction models already enforce such cross-view consistency in their latent representations, but currently apply it only to geometry.

In this work, we propose \textbf{IRIS-SLAM} (\textbf{I}ntegrated \textbf{R}econstruction with \textbf{I}nstance \textbf{S}emantics \textbf{SLAM}), a unified RGB semantic SLAM system that leverages feed-forward geo-semantic reconstruction to achieve instance-level, cross-view consistent mapping and re-localization. IRIS-SLAM treats semantic mapping and loop closure as two manifestations of the same instance-level consistency constraint, enabling robust open-vocabulary mapping and reliable re-localization under extreme viewpoint changes within a single integrated framework. Our contributions are summarized as follows:

\begin{itemize}
\item \textbf{Integrated Geo-Instance SLAM Framework}: IRIS-SLAM is the first RGB-only semantic SLAM that integrates multi-view geometry and instance-level features in a unified latent space, inherently supporting instance-level association.
\item \textbf{Unified Geo-Instance Front-end Model and Synergized Association}: We extend feed-forward reconstruction models to jointly generate dense geometry and cross-view consistent instance embeddings, allowing semantic entities to directly support data association and resolve ambiguities.

\item \textbf{Viewpoint-Invariant Loop Closure}: We present a loop closure detection method guided by joint geometric and instance-level consistency. By leveraging viewpoint-agnostic semantic anchors, our approach achieves significantly higher loop closure reliability than appearance-based methods under large viewpoint changes.
\item Comprehensive evaluations demonstrate that IRIS-SLAM achieves robust open-vocabulary mapping and reliable loop closure in cluttered scenes, surpassing state-of-the-art geometry- and appearance-based baselines.
\end{itemize}

\section{Related Work}

\subsection{Feed-forward 3D Reconstruction Models}
In recent years, feed-forward 3D reconstruction models, notably DUSt3R \cite{dust3r_cvpr24}, MASt3R \cite{mast3r_eccv24}, VGGT \cite{wang2025vggt}, and DepthAnything3 \cite{depthanything3}, have redefined the landscape of visual geometry estimation. By utilizing unified Transformer-based encoder-decoder architectures, these methods robustly reconstruct dense 2D-to-3D pointmaps and estimate camera poses directly from images. This paradigm effectively mitigates classical SLAM failure modes, such as viewpoint degeneracy and tracking loss in textureless regions. Building on these foundations, systems like MASt3R-SLAM \cite{murai2024_mast3rslam}, VGGT-SLAM \cite{maggio2025vggt-slam}, and VGGT-Long \cite{deng2025vggtlongchunkitloop} achieve high-fidelity geometric reconstruction through keyframe or chunk-based optimization, coupling 3D priors with global pose-graph refinement. Despite their robustness, these systems remain predominantly geometry-centric. They lack a deep semantic understanding of scene entities, which constrains their utility in embodied AI tasks that require reasoning about object-level properties. In contrast, IRIS-SLAM builds upon high-fidelity geometry to enable open-vocabulary semantic reconstruction, bridging the gap between geometric structure and semantic meaning within a unified framework.

\subsection{Open-Vocabulary Semantic Mapping}
The advent of vision-language models such as CLIP \cite{radford2021learningtransferablevisualmodels} has introduced unified linguistic feature spaces into 3D scene representations, facilitating incremental open-vocabulary mapping at both instance and region levels. ConceptFusion \cite{conceptfusion} and ConceptGraphs \cite{gu2024conceptgraphs} associate instances across frames by evaluating the intersection-over-union (IoU) between projected masks and 3D point-cloud bounding boxes, fusing multi-view features into semantic point clouds. OpenGraph \cite{xia2024opengraphopengraphfoundation} and HOV-SG \cite{werby23hovsg} extend this concept to large-scale environments and hierarchical scene graphs, enabling structured querying across floors and rooms. Similarly, OpenFusion \cite{jin2025openfusionopenvocabularyrealtimescene} and OVO-SLAM \cite{martins2024ovo} leverage region-level features or real-time perception to support object-centric mapping. However, most existing pipelines treat semantic mapping as a decoupled post-processing task that relies on pre-computed poses and heuristic 2D/3D association. Consequently, semantic cues are not exploited to improve data association or bolster robustness in challenging trajectories. Our work addresses this by leveraging multi-view consistent instance embeddings to actively guide instance association, thereby enhancing the coherence and accuracy of the resulting semantic maps.
\begin{figure*}[th]
    \centering
    \vspace{-6pt}
    \includegraphics[width=0.95\linewidth]{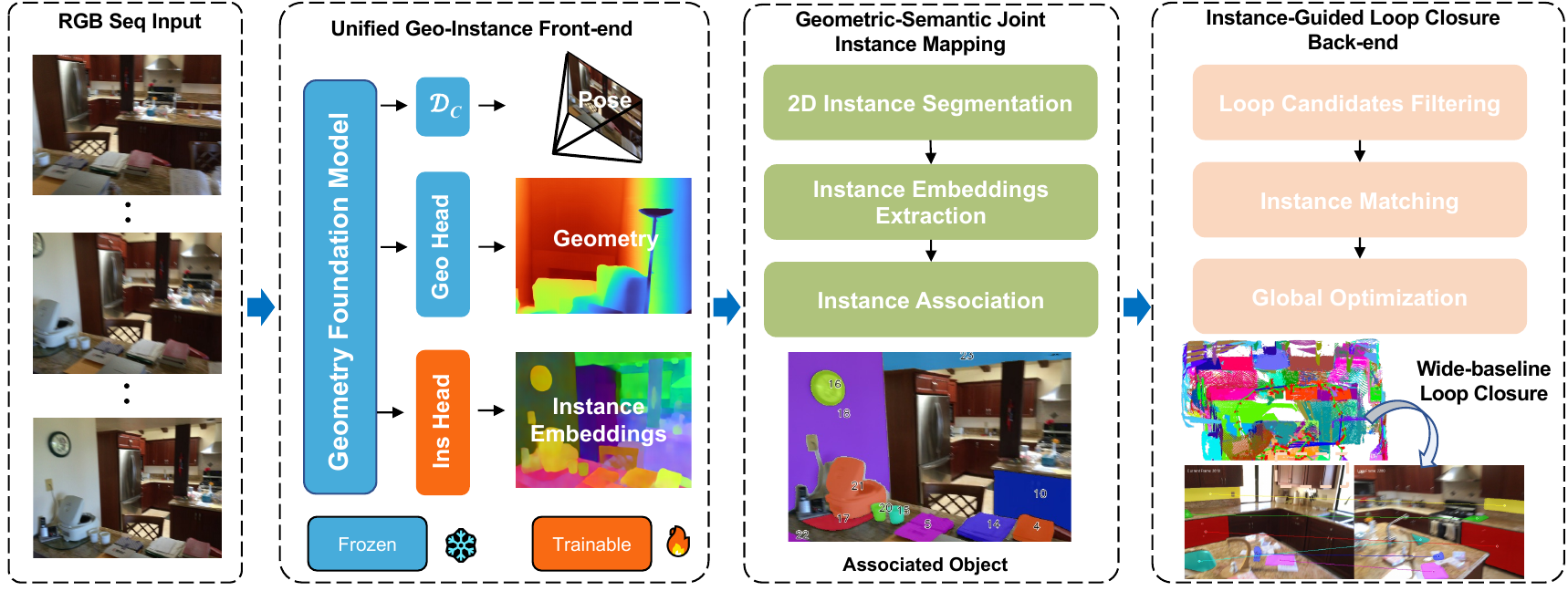}
    \caption{System Architecture of IRIS-SLAM. Our framework establishes a tight coupling between dense geometric reconstruction and instance-level understanding by extending feed-forward 3D foundation models into a shared, multi-view coherent latent space. The pipeline realizes our core contributions through three integrated modules: (1) \textbf{Unified Geo-Instance Front-end Model}: generating consistent geometric-semantic primitives from monocular streams; (2) \textbf{Instance-Grounded Association}: an instance-grounded association mechanism where instance embeddings actively drive data association; and (3) \textbf{Instance-Guided Loop Closure Back-end}: a robust loop closure module leveraging viewpoint-agnostic semantic instance anchors to maintain global map coherence even under extreme pose disparities.}
    \label{fig:system_overview}
    \vspace{-4pt}
\end{figure*}
\subsection{Loop Closure Detection in VSLAM}
Loop closure detection is essential for suppressing cumulative drift and ensuring global map consistency. Early appearance-based methods, exemplified by ORB-SLAM3 \cite{ORBSLAM3_TRO} and the VINS series \cite{vins-mono,vins-fusion}, utilize hand-crafted descriptors like ORB/BRIEF \cite{ORBfeature} and Bag-of-Words (BoW) \cite{Bow} models for retrieval. While efficient, these approaches are sensitive to illumination changes and dynamic objects due to their reliance on low-level textures. Learning-based global descriptors have significantly advanced the field; architectures such as NetVLAD \cite{arandjelović2016netvladcnnarchitectureweakly} and Patch-NetVLAD \cite{patch-netvlad} have been integrated into neural SLAM systems like Vox-Fusion++ \cite{zhai2024voxfusionvoxelbasedneuralimplicit} and Gaussian Splatting frameworks like GLC-SLAM \cite{xu2024glcslamgaussiansplattingslam} and LoopSplat \cite{zhu2025_loopsplat}. More recently, descriptors like SALAD \cite{Izquierdo_CVPR_2024_SALAD} have been adopted by feed-forward models like VGGT-SLAM \cite{maggio2025vggt-slam} to generate descriptive global vectors. 
However, these paradigms remain fundamentally appearance-centric and often fail during extreme wide-baseline scenarios. In contrast, IRIS-SLAM leverages instances with cross-view consistent features as viewpoint-agnostic semantic anchors. These anchors facilitate robust loop closure during drastic perspective shifts that would typically cause traditional descriptors to collapse.

\section{Technical Approach}
\label{sec:approach}

IRIS-SLAM integrates geometric and instance-level representations into a multi-view consistent latent space, treating data association as an intrinsic system property rather than a decoupled post-processing task (Fig.~\ref{fig:system_overview}). The pipeline begins with a Unified Geo-Instance Front-end Model (Sec.~\ref{sec:fronted}) that concurrently infers camera poses, dense depth, and high-dimensional instance embeddings from RGB sequences. These outputs drive a Geometric-Semantic Joint Instance Mapping module (Sec.~\ref{sec:seg_mapping}), where semantic entities and geometric consensus construct a persistent instance-level map integrated with SigLIP \cite{zhai2023sigmoidlosslanguageimage} features for open-vocabulary interaction. To ensure long-term topological consistency, the system concludes with an Instance-Guided Loop Closure and Global Optimization strategy (Sec.~\ref{sec:loop_opt}), which leverages clustered instance embeddings as viewpoint-agnostic anchors to eliminate cumulative drift across extreme baselines.

\subsection{Unified Geo-Instance Front-end Model}
\label{sec:fronted}

Traditional feed-forward reconstruction models typically produce purely geometric outputs, such as camera poses and depth maps, while lacking the instance-level semantic features essential for robust data association and re-localization. To address this, we propose a Unified Geo-Instance Model that integrates instance awareness directly into the geometric inference pipeline. 
Inspired by IGGT \cite{li2025iggtinstancegroundedgeometrytransformer}, we implement multi-view consistent instance embedding map prediction in a feed-forward fashion, built on the state-of-the-art feed-forward 3D reconstruction model \cite{depthanything3}. Specifically, we introduce an additional DPT-like instance head to its existing geometric head. Both heads share the feature tokens extracted from the backbone, enabling the model to jointly perform geometric reconstruction and 8-dimensional dense instance embedding prediction.

To encourage IRIS-SLAM learns cross-view consistent feature representations, we adopt the supervision as a ``pull'' and ``push'' contrastive mechanism. This design enables viewpoint-agnostic aggregation of features from the same instance across multiple views, while preserving discriminability between different instances.

The \textbf{Intra-view Pull Loss} enforces feature compactness by pulling each pixel feature $\mathbf{f}_p$ toward its corresponding per-mask feature centroid $\boldsymbol{\mu}$:
\begin{equation} 
\scriptsize
\mathcal{L}_\text{{intra\_pull}} = \frac{1}{|\mathcal{M}|} \sum_{\substack{p \in \mathcal{M}}} \max(0, m_\text{pull} - S(\mathbf{f}_p, \boldsymbol{\mu})), \boldsymbol{\mu} = \frac{1}{|\mathcal{M}|} \sum_{\substack{p \in \mathcal{M}}} \mathbf{f}_p ,
\end{equation}
where $S(\boldsymbol{u}, \boldsymbol{v})$ denotes the cosine similarity and $m_\text{pull}$ is a predefined similarity margin, and $\boldsymbol{\mu}$ represents the averaged feature embedding of the mask $\mathcal{M}$. 

To align instance representations across views, we use the \textbf{Cross-view Consistency Pull Loss}:
\begin{equation} 
\small
\mathcal{L}_\text{{cross\_pull}} = \frac{1}{|\mathcal{P}|} \sum_{\substack{\boldsymbol{\mu_i}, \boldsymbol{\mu_j} \in \mathcal{P} \\ m(\boldsymbol{\mu_i}) = m(\boldsymbol{\mu_j})}} \max(0, m_\text{pull} - S(\boldsymbol{\mu_i}, \boldsymbol{\mu_j})), 
\end{equation}
where $\mathcal{P}$ denotes the set of all per-mask centroids and $m(\cdot)$ indicates the instance identity.
The \textbf{Push Loss} encourages inter-instance separation by penalizing high similarity between per-mask centroids belonging to different instances:
\begin{equation} 
\mathcal{L}_\text{{push}} = \frac{1}{|\mathcal{P}|} \sum_{\substack{\boldsymbol{\mu_i}, \boldsymbol{\mu_j} \in \mathcal{P} \\ m(\boldsymbol{\mu_i}) \neq m(\boldsymbol{\mu_j})}} \max(0, S(\boldsymbol{\mu_i}, \boldsymbol{\mu_j}) - m_\text{push}),
\end{equation}
where $m_\text{push}$ is a predefined similarity margin.

In the online reconstruction phase, given an input RGB video chunk $\mathcal{C}_k = \{I_1, \dots, I_N\}$, the unified model $\Phi_{\text{net}}$ concurrently estimates camera poses, geometry, and semantics:
\begin{equation}
\{\mathbf{T}_i, \mathbf{D}_i, \mathbf{F}_i\}_{i=1}^N = \Phi_{\text{net}}(\mathcal{C}_k)
\end{equation}
where $\mathbf{T}_i \in SE(3)$ is the camera pose, $\mathbf{D}_i \in \mathbb{R}_{+}^{H \times W}$ is the dense depth map, and $\mathbf{F}_i \in \mathbb{R}^{H \times W \times D}$ denotes the high-dimensional instance embedding map. To resolve the scale and coordinate inconsistencies inherent in independent chunk-wise inference, we apply $Sim(3)$ alignment between adjacent chunks to maintain global trajectory coherence \cite{deng2025vggtlongchunkitloop}.

\subsection{Geometric-Semantic Joint Instance Mapping}
\label{sec:seg_mapping}

The mapping module constructs a persistent semantic map by integrating aligned poses, depth maps, and consistent instance embeddings. By incorporating vision-language priors such as SigLIP \cite{zhai2023sigmoidlosslanguageimage}, the system supports open-vocabulary queries within the reconstructed 3D space.

\noindent \textbf{Instance Segmentation and Feature Extraction.}
Instead of relying on computationally intensive external segmentors, we perform mask segmentation by exploiting the intrinsic multi-view consistency of our front-end instance embeddings. We observe that pixels belonging to the same physical entity naturally form dense clusters within our learned feature space. 

To formalize this, we define a clustering operator $\beta(\mathbf{F}_t, \epsilon)$, implemented as a similarity-based clustering algorithm on the feature map $\mathbf{F}_t$. This operator groups dense per-pixel instance features into disjoint instance masks $\mathcal{M}_t$:
\begin{equation}
\mathcal{M}_t = \{m_1, m_2, \dots, m_K\} = \beta(\mathbf{F}_t, \epsilon)
\end{equation}
where $\epsilon$ denotes the similarity threshold. Subsequently, to obtain a compact descriptor for association, we apply a \texttt{PoolAndNormalize} operation. This aggregates pixel-wise features within each mask $m_k$ via average pooling followed by $L_2$-normalization:
\begin{equation}
\mathbf{f}_k = \frac{\sum_{p \in m_k} \mathbf{F}_t(p)}{\left\|\sum_{p \in m_k} \mathbf{F}_t(p)\right\|_2}
\label{eq:sem-feat}
\end{equation}

\noindent \textbf{Joint Instance Association.}
We employ a joint association strategy to unify current observations with the global instance map. We project the existing 3D instances $\mathcal{V}_{t-1}$ into the current camera view using the pose $\mathbf{T}_{t}$ to generate projected map masks $\{\tilde{m}_j\}$. The association affinity $A(k, j)$ between a current mask $m_k$ and a global instance $j$ is a weighted fusion of geometric and instance consistency:
\begin{equation}
    A(k, j) = \alpha \cdot \mathrm{IoU}(m_k, \tilde{m}_j) + \beta \cdot \cos(\mathbf{f}_k, \mathbf{b}_j)
\end{equation}
where $\mathbf{b}_j \in \mathcal{B}_{t-1}$ represents the stored prototype in the feature database. Matches exceeding a confidence threshold $\tau_{\mathrm{match}}$ trigger an update of the feature bank using a running average strategy (denoted as \texttt{RunningAvg}) to smooth out noise; otherwise, a new instance is initialized.

\begin{algorithm}[t]
\caption{Geometric-Semantic Joint Instance Association}
\label{alg:association}
\SetAlgoLined
\DontPrintSemicolon
\SetKwInOut{Input}{Input}
\SetKwInOut{Output}{Output}

\Input{Depth $\mathbf{D}_t$, Pose $\mathbf{T}_t$, instance embedding map $\mathbf{F}_t$; \\
Instance map $\mathcal{V}_{t-1}$, Feature database $\mathcal{B}_{t-1}$}

\Output{Instance IDs $\{\mathcal{I}_k\}$ for current masks $\{m_k\}$}
\BlankLine
 Masks clustering: $\{m_k\}_{k=1}^{K} \leftarrow \beta(\mathbf{F}_t, \epsilon)$\;
Project $\mathcal{V}_{t-1}$ to current frame to get map projections $\{\tilde{m}_j\}$\;

\ForEach{current mask $m_k \in \{m_k\}_{k=1}^{K}$}{
    Extract feature: $\mathbf{f}_k \leftarrow \text{PoolAndNormalize}(\mathbf{F}_t, m_k)$\;
    Identify candidates: $\Omega_k \leftarrow \{j \mid \mathrm{IoU}(m_k, \tilde{m}_j) > 0\}$\;

    \eIf{$\Omega_k = \emptyset$}{
        Create new ID $j_{\text{new}}$; Initialize $\mathcal{B}_{t}[j_{\text{new}}] \leftarrow \mathbf{f}_k$\;
        Assign $\mathcal{I}_k \leftarrow j_{\text{new}}$\;
    }{
        \ForEach{candidate $j \in \Omega_k$}{
            $S_{\mathrm{geo}} \leftarrow \mathrm{IoU}(m_k, \tilde{m}_j)$, 
            $S_{\mathrm{sem}} \leftarrow \mathbf{f}_k^\top \cdot \mathcal{B}_{t-1}[j]$\;
            $A(k, j) \leftarrow \alpha S_{\mathrm{geo}} + \beta S_{\mathrm{sem}}$\;
        }
        
        $j^* \leftarrow \arg\max_{j \in \Omega_k} A(k, j)$\;
        
        \If{$A(k, j^*) > \tau_{\mathrm{match}}$}{
            Assign $\mathcal{I}_k \leftarrow j^*$\;
            Update bank: $\mathcal{B}_{t}[j^*] \leftarrow \text{RunningAvg}(\mathcal{B}_{t-1}[j^*], \mathbf{f}_k)$\;
        }
        \Else{
            Create new ID $j_{\text{new}}$; Initialize $\mathcal{B}_{t}[j_{\text{new}}] \leftarrow \mathbf{f}_k$\;
            Assign $\mathcal{I}_k \leftarrow j_{\text{new}}$\;
        }
    }
}
\Return $\{\mathcal{I}_k\}$\;
\end{algorithm}

\subsection{Instance-Guided Loop Closure Back-end}
\label{sec:loop_opt}



The loop closure stage transitions SLAM from incremental data association to global scene recognition. Most existing approaches rely either on global appearance similarity or purely geometric consistency. Appearance-based representations focus on image-level texture or semantic cues while ignoring fine-grained geometric and instance-level constraints, requiring substantial viewpoint overlap for reliable matching. In contrast, geometry-only validation lacks semantic cues to filter false positives. As a result, features stored in the \textbf{Feature Database} ($\mathcal{B}$), which is constructed incrementally within local spatiotemporal windows, become unreliable for robust loop closure over long trajectories.

To mitigate this issue, we introduce a hierarchical verification pipeline based on \textbf{contextual re-inference}. We first identify a set of loop candidates $\mathcal{K}{\text{cand}}$ by retrieving keyframes within a spatial neighborhood of the current pose estimate $\mathbf{T}t$. Rather than relying on potentially drifted features stored in $\mathcal{B}$, we re-evaluate each candidate at the image level: for every $I_k \in \mathcal{K}{\text{cand}}$, we form an image pair $\mathcal{C}{\text{loop}} = {I_k, I_t}$ and jointly process it using the front-end $\Phi_{\text{net}}$.

This joint inference projects both candidates into a shared latent space, effectively re-aligning their representations and producing synchronized instance embeddings ${\mathbf{F}_k, \mathbf{F}t}$. We then perform explicit instance-level matching between the two frames. A loop closure is accepted only if the number of mutually consistent instances matches-verified through both semantic similarity and geometric agreement—exceeds a threshold $\tau_{\text{loop}}$.

\noindent \textbf{Global Optimization.} Upon loop validation, we establish a cross-chunk $\mathrm{Sim}(3)$ constraint to rectify scale and trajectory drift. The global state $\mathcal{X}$ is refined by minimizing the following joint cost function:
\begin{equation}
\mathcal{X}^* = \arg \min_{\mathcal{X}} \left( \sum_{i} \rho ( \| \mathbf{e}_{i,i+1} \|^2_{\Sigma} ) + \sum_{(j,k) \in \mathcal{L}} \rho ( \| \mathbf{e}_{j,k} \|^2_{\Sigma} ) \right)
\end{equation}
where $\mathbf{e}_{j,k} = \log( \mathbf{S}_{jk}^{-1} \mathbf{S}_{k} \mathbf{S}_{j}^{-1} )$ represents the $\mathrm{Sim}(3)$ residual between chunks $j$ and $k$, and $\rho$ denotes a robust Huber kernel.
\section{Experiments}

\noindent \textbf{Implementation Details.}
We implement IRIS-SLAM using PyTorch and execute all experiments on a workstation equipped with an NVIDIA RTX 4090 GPU to ensure real-time processing capabilities. Following the protocols established by VGGT-SLAM \cite{maggio2025vggt-slam} and VGGT-Long \cite{deng2025vggtlongchunkitloop}, our front-end processes RGB streams in chunks of 120 frames with a 60-frame overlap between consecutive windows to maintain temporal continuity. For semantic understanding, we integrate the SigLIP \cite{zhai2023sigmoidlosslanguageimage} model to facilitate open-vocabulary interaction. To train the Unified Geo-Instance front-end Model, we initialize the architecture with Depth-Anything-v3-Giant weights. We freeze the backbone parameters and focus optimization exclusively on the proposed instance head. Training is conducted on the InsScene-15K dataset \cite{li2025iggtinstancegroundedgeometrytransformer} using 8 NVIDIA L20 GPUs.

\noindent \textbf{Baselines.} To evaluate the flexibility and robustness of our framework, we configure IRIS under two complementary modes: \textbf{IRIS-Mapping}, which utilizes Ground-Truth (GT) trajectories and depth maps to establish the upper-bound performance of our semantic reconstruction module; and \textbf{IRIS-SLAM}, the complete system utilizing chunk-aligned pose and depth predictions from our front-end. We benchmark our method across three critical dimensions: camera pose estimation, open-vocabulary semantic mapping, and loop closure efficacy. For pose estimation, we compare IRIS-SLAM against a representative spectrum of traditional and learning-based systems, including ORB-SLAM3 \cite{ORBSLAM3_TRO}, the flow-based DROID-SLAM \cite{teed2022droidslamdeepvisualslam}, and SOTA feed-forward frameworks such as Mast3r-SLAM \cite{murai2024_mast3rslam} and the VGGT series \cite{maggio2025vggt-slam, deng2025vggtlongchunkitloop}. For semantic mapping, we evaluate IRIS against leading 3D Gaussian Splatting approaches, including \textit{Open-Gaussian} \cite{wu2024opengaussian} and \textit{LangSplat} \cite{langsplat}, alongside incremental methods like \textit{ConceptFusion} \cite{conceptfusion} and \textit{ConceptGraphs} \cite{gu2024conceptgraphs}. Finally, for loop closure, we validate our instance-guided detection against established descriptors, including the DBoW vocabulary in ORB-SLAM3 \cite{ORBfeature, Bow}, NetVLAD \cite{arandjelović2016netvladcnnarchitectureweakly}, and the SALAD \cite{Izquierdo_CVPR_2024_SALAD} model.

\noindent \textbf{Datasets.} To rigorously evaluate camera pose estimation, we utilize nine sequences from the TUM RGB-D dataset \cite{TUMD}, covering various motion patterns and environmental conditions. For open-vocabulary semantic mapping, we benchmark IRIS on both synthetic and real-world datasets: six synthetic scenes from Replica \cite{replica19arxiv} (office0-4 and room0-2) and a eight-scene subset of ScanNet \cite{dai2017scannet} (0000, 0059, 0106, 0169, 0181, and 0207). These ScanNet sequences represent a broad spectrum of indoor environments characterized by complex geometries and varying clutter. For loop closure, we follow the methodology in \cite{10935649} to construct a dedicated place-recognition benchmark from the ScanNet subset. We sample place and query frames from sequence keyframes at a five-frame interval. A keyframe is designated as a place frame if its spatial overlap with existing frames falls below 0.3; otherwise, it is categorized as a query frame for retrieval assessment.

\begin{table*}[ht]
\vspace{-4pt}
\scriptsize
\setlength{\tabcolsep}{5.5pt} 
\centering
\caption{Camera pose estimation comparison with RMSE$\downarrow$ on TUM RGB-D \cite{TUMD} (unit: m). Colors are normalized \textbf{column-wise} from \colorbox{green1}{green} (best in column) to \colorbox{red2}{red} (worst in column).}
\label{tab:trajcompare}
\begin{tabular}{lcccccccccc}
\toprule
\textbf{Methods} & \textbf{AVG} & \textbf{360} & \textbf{desk} & \textbf{desk2} & \textbf{floor} & \textbf{plant} & \textbf{room} & \textbf{rpy} & \textbf{teddy} & \textbf{xyz} \\
\midrule
ORBSLAM3 & \xmark & \xmark & \getcol{0.032}{0.026}{0.058}0.017 & \getcol{0.091}{0.032}{0.111}0.021 & \xmark & \getcol{0.045}{0.022}{0.085}0.034 & \xmark & \xmark & \xmark & \getcol{0.012}{0.012}{0.099}0.009 \\
\hline
Droid-SLAM & \getcol{0.163}{0.049}{0.163}0.163 & \getcol{0.202}{0.059}{0.202}0.202 & \getcol{0.032}{0.026}{0.058}0.032 & \getcol{0.091}{0.032}{0.111}0.091 & \getcol{0.064}{0.055}{0.254}0.064 & \getcol{0.045}{0.022}{0.085}0.045 & \getcol{0.918}{0.057}{0.918}0.918 & \getcol{0.056}{0.034}{0.206}0.056 & \getcol{0.045}{0.032}{0.126}0.045 & \getcol{0.012}{0.012}{0.099}0.012 \\
\hline
Mast3r-SLAM & \getcol{0.060}{0.049}{0.163}0.060 & \getcol{0.070}{0.059}{0.202}0.070 & \getcol{0.035}{0.026}{0.058}0.035 & \getcol{0.055}{0.032}{0.111}0.055 & \getcol{0.056}{0.055}{0.254}0.056 & \getcol{0.035}{0.022}{0.085}0.035 & \getcol{0.118}{0.057}{0.918}0.118 & \getcol{0.041}{0.034}{0.206}0.041 & \getcol{0.114}{0.032}{0.126}0.114 & \getcol{0.020}{0.012}{0.099}0.020 \\
\hline
VGGT-SLAM & \getcol{0.074}{0.049}{0.163}0.074 & \getcol{0.123}{0.059}{0.202}0.123 & \getcol{0.040}{0.026}{0.058}0.040 & \getcol{0.055}{0.032}{0.111}0.055 & \getcol{0.254}{0.055}{0.254}0.254 & \getcol{0.022}{0.022}{0.085}0.022 & \getcol{0.088}{0.057}{0.918}0.088 & \getcol{0.041}{0.034}{0.206}0.041 & \getcol{0.032}{0.032}{0.126}0.032 & \getcol{0.016}{0.012}{0.099}0.016 \\
\hline
VGGT-Long & \getcol{0.110}{0.049}{0.163}0.110 & \getcol{0.118}{0.059}{0.202}0.118 & \getcol{0.058}{0.026}{0.058}0.058 & \getcol{0.111}{0.032}{0.111}0.111 & \getcol{0.118}{0.055}{0.254}0.118 & \getcol{0.071}{0.022}{0.085}0.071 & \getcol{0.155}{0.057}{0.918}0.155 & \getcol{0.140}{0.034}{0.206}0.140 & \getcol{0.120}{0.032}{0.126}0.120 & \getcol{0.099}{0.012}{0.099}0.099 \\
\hline
Pi-long & \getcol{0.094}{0.049}{0.163}0.094 & \getcol{0.115}{0.059}{0.202}0.115 & \getcol{0.047}{0.026}{0.058}0.047 & \getcol{0.052}{0.032}{0.111}0.052 & \getcol{0.160}{0.055}{0.254}0.160 & \getcol{0.085}{0.022}{0.085}0.085 & \getcol{0.114}{0.057}{0.918}0.114 & \getcol{0.143}{0.034}{0.206}0.143 & \getcol{0.081}{0.032}{0.126}0.081 & \getcol{0.052}{0.012}{0.099}0.052 \\
\hline
DA3-Stream & \getcol{0.087}{0.049}{0.163}0.087 & \getcol{0.059}{0.059}{0.202}0.059 & \getcol{0.034}{0.026}{0.058}0.034 & \getcol{0.042}{0.032}{0.111}0.042 & \getcol{0.107}{0.055}{0.254}0.107 & \getcol{0.060}{0.022}{0.085}0.060 & \getcol{0.105}{0.057}{0.918}0.105 & \getcol{0.206}{0.034}{0.206}0.206 & \getcol{0.126}{0.032}{0.126}0.126 & \getcol{0.044}{0.012}{0.099}0.044 \\
\hline
Ours (w/o sample) & \getcol{0.061}{0.049}{0.163}0.061 & \getcol{0.082}{0.059}{0.202}0.082 & \getcol{0.034}{0.026}{0.058}0.034 & \getcol{0.040}{0.032}{0.111}0.040 & \getcol{0.084}{0.055}{0.254}0.084 & \getcol{0.063}{0.022}{0.085}0.063 & \getcol{0.057}{0.057}{0.918}0.057 & \getcol{0.041}{0.034}{0.206}0.041 & \getcol{0.120}{0.032}{0.126}0.120 & \getcol{0.033}{0.012}{0.099}0.033 \\
\textbf{Ours (w sample)} & \getcol{0.049}{0.049}{0.163}\textbf{0.049} & \getcol{0.092}{0.059}{0.202}0.092 & \getcol{0.026}{0.026}{0.058}\textbf{0.026} & \getcol{0.032}{0.032}{0.111}\textbf{0.032} & \getcol{0.055}{0.055}{0.254}\textbf{0.055} & \getcol{0.047}{0.022}{0.085}0.047 & \getcol{0.065}{0.057}{0.918}0.065 & \getcol{0.034}{0.034}{0.206}\textbf{0.034} & \getcol{0.066}{0.032}{0.126}0.066 & \getcol{0.020}{0.012}{0.099}0.020 \\
\bottomrule
\end{tabular}
\vspace{-4pt}
\end{table*}

\begin{table*}[ht]
\centering
\caption{Zero-shot semantic evaluations on ScanNet and Replica datasets. Metrics are normalized \textbf{column-wise} from \colorbox{green1}{green} (best) to \colorbox{red2}{red} (worst). IRIS-SLAM achieves competitive results even with estimated (Est/Est) poses and depth.}
\label{tab:semantic_eval_combined}
\begin{subtable}{0.49\textwidth}
    \centering
    \caption{ScanNet dataset}
    \resizebox{\columnwidth}{!}{%
    \setlength{\tabcolsep}{4pt}
    \begin{tabular}{lc c cccc}
    \toprule
    \textbf{Method} & \textbf{Online} & \textbf{Pose/Depth} & \textbf{mIoU}$\uparrow$ & \textbf{mAcc}$\uparrow$ & \textbf{f-mIoU}$\uparrow$ & \textbf{f-Acc}$\uparrow$ \\
    \midrule
    Open-Gaussian & \textcolor{red}{\xmark} & GT/GT & \getsem{08.64}{06.69}{39.93}08.64 & \getsem{17.86}{14.89}{55.36}17.86 & \getsem{23.71}{16.62}{53.62}23.71 & \getsem{26.44}{19.57}{66.78}26.44 \\
    Lang-Splat & \textcolor{red}{\xmark} & GT/GT & \getsem{07.22}{06.69}{39.93}07.22 & \getsem{21.01}{14.89}{55.36}21.01 & \getsem{27.59}{16.62}{53.62}27.59 & \getsem{32.21}{19.57}{66.78}32.21 \\
    Grasp-Splats & \textcolor{green}{\cmark} & GT/GT & \getsem{06.69}{06.69}{39.93}06.69 & \getsem{14.89}{14.89}{55.36}14.89 & \getsem{16.62}{16.62}{53.62}16.62 & \getsem{19.57}{19.57}{66.78}19.57 \\
    Omni-Map & \textcolor{green}{\cmark} & GT/GT & \getsem{25.42}{06.69}{39.93}25.42 & \getsem{50.93}{14.89}{55.36}50.93 & \getsem{50.86}{16.62}{53.62}50.86 & \getsem{57.05}{19.57}{66.78}57.05 \\
    HOV-SG & \textcolor{red}{\xmark} & GT/GT & \getsem{20.76}{06.69}{39.93}20.76 & \getsem{41.50}{14.89}{55.36}41.50 & \getsem{38.34}{16.62}{53.62}38.34 & \getsem{45.50}{19.57}{66.78}45.50 \\
    Concept-Fusion & \textcolor{green}{\cmark} & GT/GT & \getsem{08.50}{06.69}{39.93}08.50 & \getsem{31.81}{14.89}{55.36}31.81 & \getsem{30.05}{16.62}{53.62}30.05 & \getsem{36.64}{19.57}{66.78}36.64 \\
    Concept-Graph & \textcolor{green}{\cmark} & GT/GT & \getsem{16.29}{06.69}{39.93}16.29 & \getsem{34.07}{14.89}{55.36}34.07 & \getsem{33.29}{16.62}{53.62}33.29 & \getsem{41.60}{19.57}{66.78}41.60 \\
    Open-Fusion & \textcolor{green}{\cmark} & GT/GT & \getsem{18.02}{06.69}{39.93}18.02 & \getsem{44.31}{14.89}{55.36}44.31 & \getsem{43.82}{16.62}{53.62}43.82 & \getsem{51.09}{19.57}{66.78}51.09 \\
    OVO-mapping & \textcolor{green}{\cmark} & GT/GT & \getsem{31.58}{06.69}{39.93}31.58 & \getsem{45.54}{14.89}{55.36}45.54 & \getsem{47.43}{16.62}{53.62}47.43 & \getsem{61.58}{19.57}{66.78}61.58 \\
    \midrule
    \textbf{IRIS-mapping} & \textcolor{green}{\cmark} & GT/GT & \getsem{39.93}{06.69}{39.93}\textbf{39.93} & \getsem{55.36}{14.89}{55.36}\textbf{55.36} & \getsem{53.62}{16.62}{53.62}\textbf{53.62} & \getsem{66.78}{19.57}{66.78}\textbf{66.78} \\
    \textbf{IRIS-SLAM} & \textcolor{green}{\cmark} & \textbf{Est/Est} & \getsem{31.62}{06.69}{39.93}31.62 & \getsem{45.08}{14.89}{55.36}45.08 & \getsem{43.85}{16.62}{53.62}43.85 & \getsem{58.32}{19.57}{66.78}58.32 \\
    \bottomrule
    \end{tabular}%
    }
\end{subtable}
\hfill
\begin{subtable}{0.49\textwidth}
    \centering
    \caption{Replica dataset}
    \resizebox{\columnwidth}{!}{%
    \setlength{\tabcolsep}{4pt}
    \begin{tabular}{l c c c c c c}
    \toprule
    \textbf{Method} & \textbf{Online} & \textbf{Pose/Depth} & \textbf{mIoU}$\uparrow$ & \textbf{mAcc}$\uparrow$ & \textbf{f-mIoU}$\uparrow$ & \textbf{f-Acc}$\uparrow$ \\
    \midrule
    Open-Gaussian & \textcolor{red}{\xmark} & GT/GT & \getsem{06.82}{04.75}{31.15}06.82 & \getsem{16.66}{16.66}{44.54}16.66 & \getsem{15.41}{15.41}{64.42}15.41 & \getsem{18.08}{18.08}{72.30}18.08 \\
    Lang-Splat & \textcolor{red}{\xmark} & GT/GT & \getsem{10.00}{04.75}{31.15}10.00 & \getsem{22.93}{16.66}{44.54}22.93 & \getsem{39.69}{15.41}{64.42}39.69 & \getsem{44.16}{18.08}{72.30}44.16 \\
    Grasp-Splats & \textcolor{green}{\cmark} & GT/GT & \getsem{10.42}{04.75}{31.15}10.42 & \getsem{23.79}{16.66}{44.54}23.79 & \getsem{42.67}{15.41}{64.42}42.67 & \getsem{52.39}{18.08}{72.30}52.39 \\
    Omni-Map & \textcolor{green}{\cmark} & GT/GT & \getsem{29.06}{04.75}{31.15}29.06 & \getsem{44.54}{16.66}{44.54}\textbf{44.54} & \getsem{64.42}{15.41}{64.42}\textbf{64.42} & \getsem{72.22}{18.08}{72.30}72.22 \\
    HOV-SG & \textcolor{red}{\xmark} & GT/GT & \getsem{23.79}{04.75}{31.15}23.79 & \getsem{39.59}{16.66}{44.54}39.59 & \getsem{48.86}{15.41}{64.42}48.86 & \getsem{55.15}{18.08}{72.30}55.15 \\
    Concept-Fusion & \textcolor{green}{\cmark} & GT/GT & \getsem{04.75}{04.75}{31.15}04.75 & \getsem{19.29}{16.66}{44.54}19.29 & \getsem{25.30}{15.41}{64.42}25.30 & \getsem{28.99}{18.08}{72.30}28.99 \\
    Concept-Graph & \textcolor{green}{\cmark} & GT/GT & \getsem{16.46}{04.75}{31.15}16.46 & \getsem{31.51}{16.66}{44.54}31.51 & \getsem{35.69}{15.41}{64.42}35.69 & \getsem{42.44}{18.08}{72.30}42.44 \\
    Open-Fusion & \textcolor{green}{\cmark} & GT/GT & \getsem{16.37}{04.75}{31.15}16.37 & \getsem{35.15}{16.66}{44.54}35.15 & \getsem{51.65}{15.41}{64.42}51.65 & \getsem{60.37}{18.08}{72.30}60.37 \\
    OVO-SLAM & \textcolor{green}{\cmark} & GT/GT & \getsem{29.93}{04.75}{31.15}29.93 & \getsem{43.60}{16.66}{44.54}43.60 & \getsem{57.00}{15.41}{64.42}57.00 & \getsem{65.39}{18.08}{72.30}65.39 \\
    \midrule
    \textbf{IRIS-mapping} & \textcolor{green}{\cmark} & GT/GT & \getsem{31.15}{04.75}{31.15}\textbf{31.15} & \getsem{40.63}{16.66}{44.54}40.63 & \getsem{60.00}{15.41}{64.42}60.00 & \getsem{72.30}{18.08}{72.30}\textbf{72.30} \\
    \textbf{IRIS-SLAM} & \textcolor{green}{\cmark} & \textbf{Est/Est} & \getsem{20.22}{04.75}{31.15}20.22 & \getsem{29.43}{16.66}{44.54}29.43 & \getsem{50.28}{15.41}{64.42}50.28 & \getsem{65.00}{18.08}{72.30}65.00 \\
    \bottomrule
    \end{tabular}%
    }
\end{subtable}
\vspace{-6pt}
\end{table*}

\noindent \textbf{Metrics.} To quantify pose accuracy, we report the root mean square error (RMSE) of the absolute trajectory error (ATE) in meters. For open-vocabulary understanding, we employ four standard metrics: mean Intersection-over-Union (mIoU), frequency-weighted IoU (f-mIoU), mean Accuracy (mAcc), and frequency-weighted Accuracy (f-Acc). Adhering to the protocol for 3D multi-class classification, we utilize GT semantic labels as query strings. For each map element, the optimal category is determined by the maximum cosine similarity between its embedded feature and the candidate label embeddings. This mapping translates the continuous open-vocabulary feature space into a discrete label space for objective benchmarking. 
Regarding loop closure, we evaluate retrieval performance using precision, recall, and F1 score based on top-1 retrieval results. A loop candidate is considered a true positive if its spatial overlap with the corresponding place frame exceeds a predefined threshold ratio.

\subsection{Camera Pose Estimation} 
\label{sec:exp_pose}

The trajectory accuracy comparisons are summarized in Table \ref{tab:trajcompare}. While DA3-Streaming \cite{depthanything3} is closely related to our geometric module, IRIS-SLAM fundamentally differs in its global optimization strategy. Specifically, instead of relying on traditional appearance-based descriptors for loop detection, we employ an instance-guided mechanism that identifies a substantially higher volume of valid loop closures. This approach generates a denser set of inter-chunk constraints, significantly enhancing global trajectory consistency.

Notably, we observe that temporal downsampling of input images further refines trajectory accuracy. This improvement stems from the reduction in the total number of processed chunks, which effectively minimizes the cumulative drift incurred during the sequential chunk-alignment process. As shown in Table \ref{tab:trajcompare}, IRIS-SLAM achieves SOTA performance across the majority of TUM RGB-D sequences.

\subsection{3D Semantic Mapping}
\label{sec:exp_seg_mapping}

Building upon high-precision geometric pose estimation, IRIS-SLAM integrates instance-level features to achieve robust and consistent cross-view association. We benchmark it on the ScanNet and Replica datasets, with the zero-shot semantic mapping results summarized in Tables \ref{tab:semantic_eval_combined}. IRIS-Mapping/SLAM consistently outperforms SOTA baselines across all evaluated semantic metrics, a success we primarily attribute to our joint geometric-semantic association strategy.

Traditional instance association frameworks rely heavily on 2D or 3D Intersection-over-Union (IoU), which fundamentally necessitates physical spatial overlap between consecutive observations. Such geometric constraints frequently fail ($IoU \approx 0$) under drastic viewpoint changes or significant occlusions, leading to association failures and fragmented map representations. In contrast, IRIS-SLAM leverages viewpoint-agnostic instance embeddings as stable semantic anchors. This allows the system to maintain robust data association even when geometric co-visibility is negligible, ensuring superior global semantic map consistency.

Furthermore, IRIS-SLAM represents a transition toward a more unified and elegant SLAM architecture. Unlike existing baselines that require pre-computed GT poses and depth maps, our framework operates as a cohesive, end-to-end streaming process. Notably, even when utilizing exclusively estimated (Est/Est) poses and depth from our front-end, IRIS-SLAM achieves semantic mapping accuracy that surpasses most baselines relying on GT geometry. This empirical evidence shows that our unified pipeline both simplifies the SLAM architecture and achieves state-of-the-art performance without relying on external geometric priors, such as sensor depth measurements.

\begin{figure*}[ht]
\vspace{-4pt}
\centering


\begin{subfigure}{0.49\linewidth}
\centering
\includegraphics[width=\linewidth]{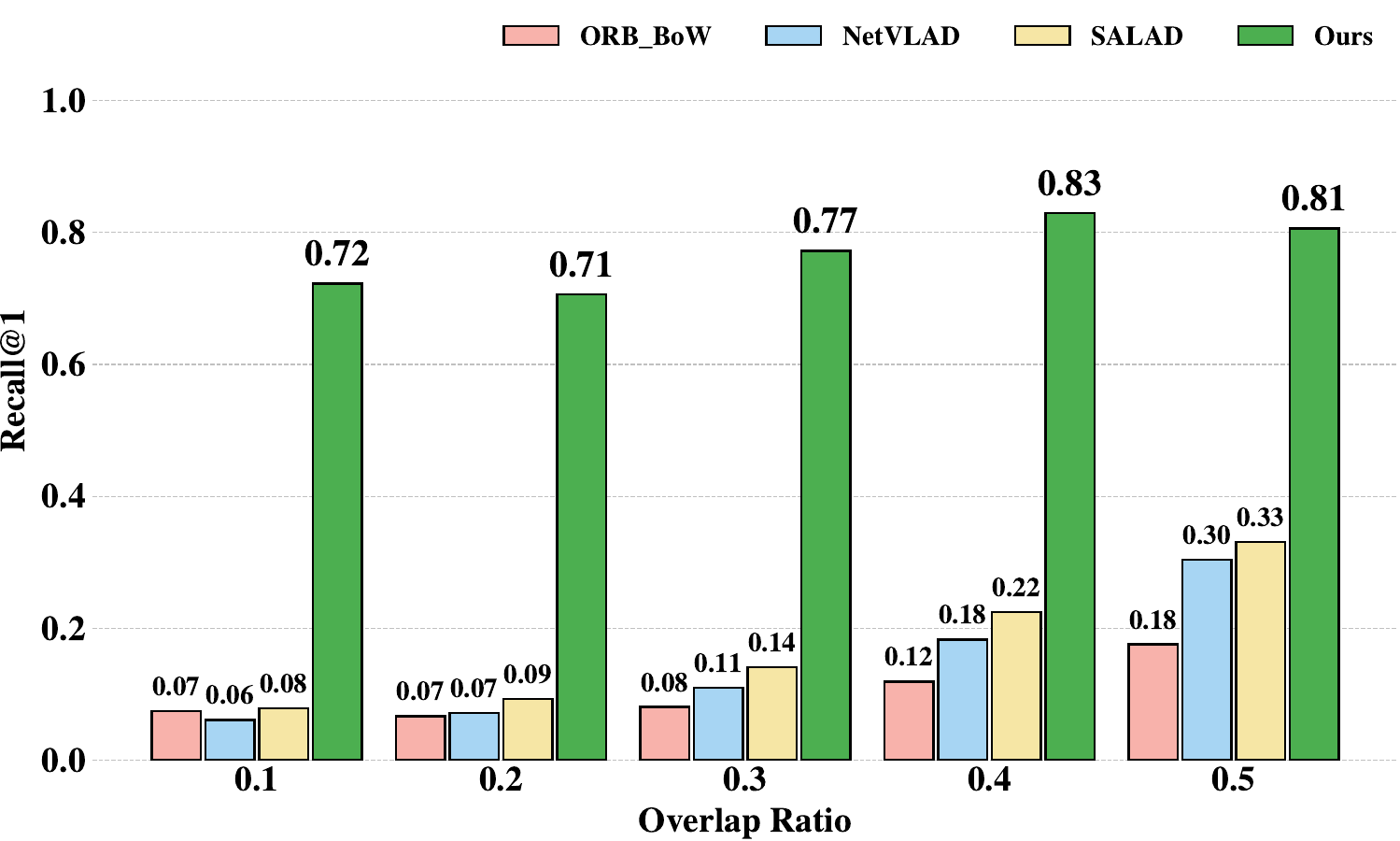}
\caption{Recall@1}
\label{fig:recall}
\end{subfigure}
\hfill
\begin{subfigure}{0.49\linewidth}
\centering
\includegraphics[width=\linewidth]{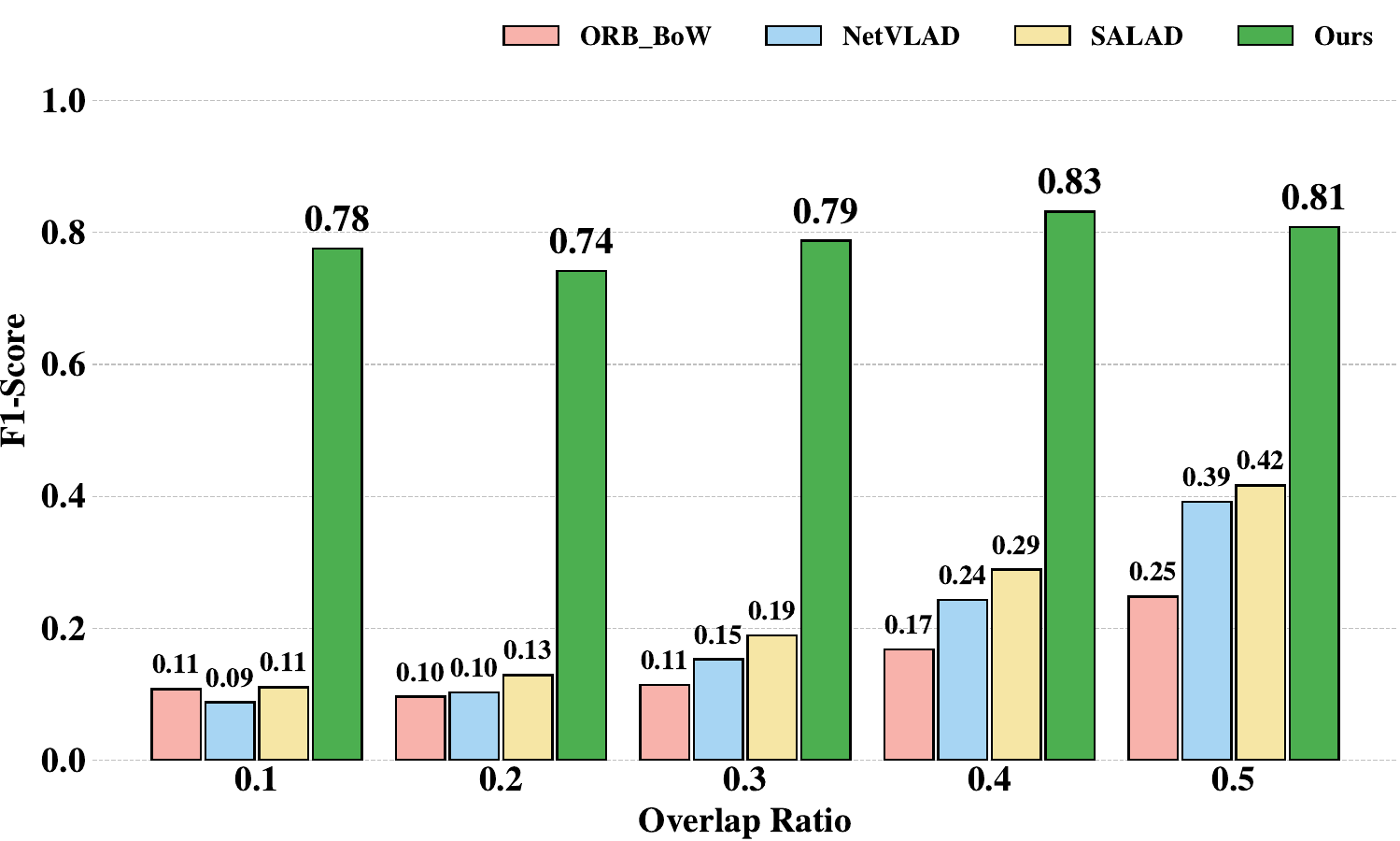}
\caption{F1-Score}
\label{fig:f1}
\end{subfigure}
\caption{Comparative loop closure performance across different overlap thresholds $\tau$. IRIS-SLAM outperforms baselines with significant margins, especially in wide-baseline scenarios ($\tau=0.1$) where traditional descriptors like NetVLAD and ORB\_BoW collapse. Our method maintains a resilient F1-score$\uparrow$ and high Recall@1$\uparrow$ by leveraging instance and structural consistency.}
\label{fig:loop_closure_results}
\end{figure*}

\subsection{Instance-guided Loop Closure}
\label{sec:exp_loop}

To evaluate loop closure reliability, we define GT matches based on the spatial overlap ratio between query and place frames. A pair is labeled positive if their shared field-of-view (FoV) exceeds a threshold $\tau \in \{0.1, 0.2, 0.3, 0.4, 0.5\}$. By decreasing $\tau$, we simulate "wide-baseline" scenarios that represent a known failure zone for conventional SLAM, where sparse visual overlap renders low-level geometric features and holistic descriptors insufficient.



As illustrated in Fig.~\ref{fig:loop_closure_results}, IRIS-SLAM outperforms all baselines by a significant margin. Specifically, at the lower bound of visual overlap ($\tau = 0.1$), where holistic descriptors such as NetVLAD and ORB-BoW collapse (F1$<$0.15) due to feature sparsity, our method maintains a resilient F1-score of 0.78 by leveraging instance-level structural consistency. At $\tau = 0.5$, our method achieves a Recall@1 of 0.81 ($4.5\times$ higher than NetVLAD), validating that instance-level signatures offer superior discriminative power over conventional aggregated global features.

The cross-view robustness of our approach is further quantified in Table ~\ref{tab:crossview}. While competitive baselines like SALAD and NetVLAD perform adequately under small angular displacements ($0$-$30^\circ$), their precision and recall deteriorate sharply as the viewpoint angle increases. Notably, in the challenging $30$-$60^\circ$ cross-view range, our loop closure method retains a remarkably high Precision of \textbf{0.297} and an F1-score of \textbf{0.228}, whereas SALAD fails to retrieve any valid matches. Our F1-score is more than $7\times$ higher than that of the most resilient baseline, ORB-BoW (0.031).

\begin{figure}[t]
    \centering
    \includegraphics[width=1.0\linewidth]{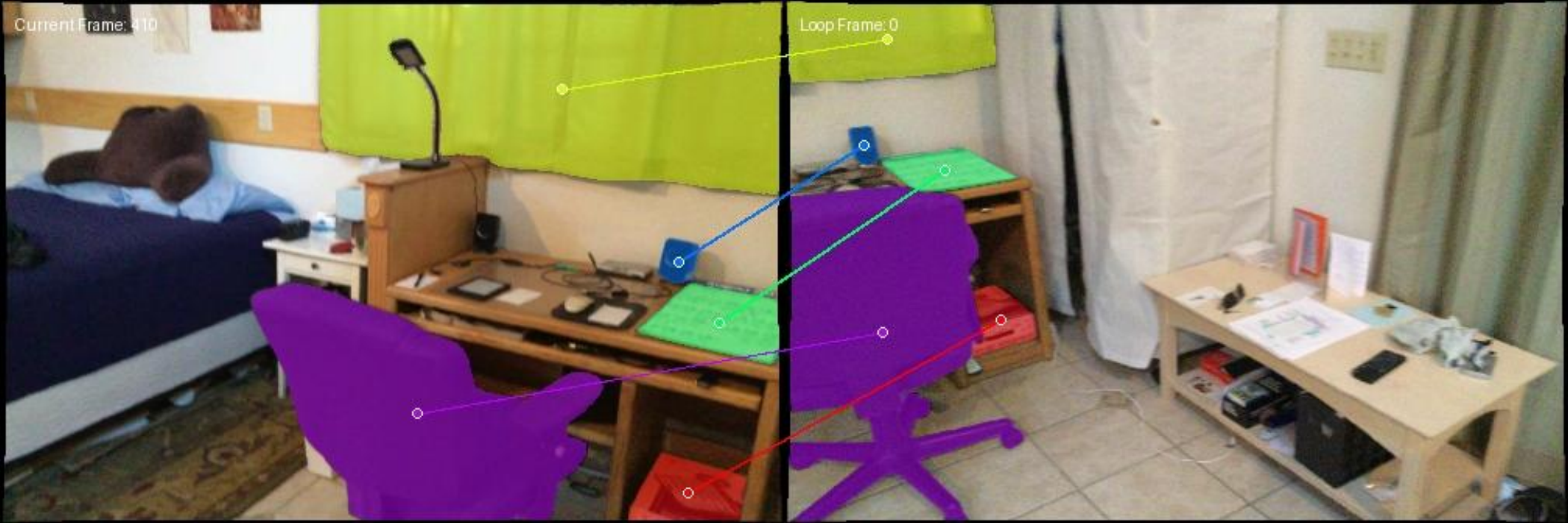}
    \caption{A loop pair with a \textbf{$67.4^{\circ}$} viewpoint shift. While ours correctly identifies the loop, the best-performing baselines, NetVLAD and SALAD, yield scores only \textbf{$0.4596$} and \textbf{$0.2567$}.}
    \label{fig:qualitative_match}
    \vspace{-6pt}
\end{figure}

To complement the quantitative results, we present a qualitative comparison on a challenging loop closure pair with a significant viewpoint variation of $67.4^{\circ}$ (Fig.~\ref{fig:qualitative_match}), which exceeds the most difficult range ($30$-$60^{\circ}$) in our statistical evaluation. 

\begin{table}[t]
\centering
\caption{Quantitative Evaluation of Cross-View Loop Closure Performance under Increasing Viewpoint Differences.}
\label{tab:crossview}

\vspace{-2mm}
\footnotesize
\setlength{\tabcolsep}{5pt}
\begin{tabular}{lcccccc}
\toprule
\multirow{2}{*}{\textbf{Method}} & \multicolumn{2}{c}{\textbf{Precision}$\uparrow$} & \multicolumn{2}{c}{\textbf{Recall@1}$\uparrow$} & \multicolumn{2}{c}{\textbf{F1-score}$\uparrow$} \\
\cmidrule(lr){2-3} \cmidrule(lr){4-5} \cmidrule(lr){6-7}
& $0$-$30^\circ$ & $30$-$60^\circ$ & $0$-$30^\circ$ & $30$-$60^\circ$ & $0$-$30^\circ$ & $30$-$60^\circ$ \\
\midrule
NetVLAD & 0.219 & 0.018 & 0.094 & 0.007 & 0.132 & 0.010 \\
ORB+BoW & 0.170 & 0.051 & 0.071 & 0.022 & 0.100 & 0.031 \\
SALAD   & 0.270 & 0.000 & 0.130 & 0.000 & 0.175 & 0.000 \\
\textbf{Ours} & \textbf{0.844} & \textbf{0.297} & \textbf{0.617} & \textbf{0.185} & \textbf{0.713} & \textbf{0.228} \\
\bottomrule
\end{tabular}
\end{table}

In summary, IRIS-SLAM not only achieves state-of-the-art performance in standard scenarios but also demonstrates exceptional resilience in extreme cross-view cases. While image pixels vary drastically with viewpoint, the semantic objects and their relative spatial layouts remain inherently consistent. By integrating instance-level spatial verification, our loop closure method leverages this stable semantics to maintain a consistent scene representation, ensuring reliable loop detection even under substantial rotations or minimal visual overlap that typically exceed the capabilities of conventional visual features.

\subsection{Parameter Sensitivity and Ablations Analysis}
\label{sec:exp_ablations}

Table \ref{tab:sensitivity_sampling} evaluates the critical impact of geometric association and temporal sampling sparsity on mapping quality. The \textbf{w/o geo} variant, which relies exclusively on instance embedding similarity for association, exhibits a marked performance degradation across all metrics. This drop confirms that the integration of the geometric score $S_{\mathrm{geo}}$ is indispensable for resolving data association ambiguities, particularly when distinguishing between spatially adjacent instances of the same semantic category.
Furthermore, IRIS-SLAM demonstrates exceptional stability across varying temporal sampling rates. Even as the sampling interval triples (Rate 10 to 30), the system maintains competitive accuracy, consistently outperforming the dense baseline (OVO-mapping). This robustness, stemming from the synergy of coherent instance embeddings and rigid geometric constraints, ensures reliable fusion even under sparse-frame conditions.


\begin{table}[t]
\vspace{-4mm}
\centering
\caption{Sampling Rate Sensitivity and $S_{\mathrm{geo}}$ Ablations Analysis on ScanNet.}
\label{tab:sensitivity_sampling}
\scriptsize
\setlength{\tabcolsep}{6pt}
\begin{tabular}{l@{\hspace{6pt}}c@{\hspace{6pt}}cccc}
\toprule
\textbf{Method} & \textbf{Rate} & \textbf{mIoU}$\uparrow$ & \textbf{mAcc}$\uparrow$ & \textbf{f-mIoU}$\uparrow$ & \textbf{f-Acc}$\uparrow$ \\
\midrule
OVO-mapping & 10 & 31.58 & 45.54 & 47.43 & 61.58 \\
Ours (w/o geo) & 10 & 34.75 & 49.47 & 48.80 & 61.46 \\
\midrule
\multirow{3}{*}{Ours (full)} & 10 & \textbf{39.93} & \textbf{55.36} & \textbf{53.62} & \textbf{66.78} \\
 & 20 & 36.88 & 51.77 & 51.77 & 65.20 \\
 & 30 & 33.46 & 46.00 & 48.52 & 60.23 \\
\bottomrule
\end{tabular}
\vspace{-4mm}
\end{table}

\section{Conclusion} 
\label{sec:conclusion}
We presented IRIS-SLAM, a semantic SLAM system bridging dense geometric reconstruction and instance-level understanding. By extending a geometry foundation model to infer cross-view coherent instance embeddings, we treat semantic association as an intrinsic geometric property rather than a decoupled task. Evaluations on ScanNet and Replica demonstrate that our semantic-synergized association and instance-guided loop closure significantly enhance map consistency and wide-baseline retrieval. While IRIS-SLAM achieves state-of-the-art performance in semantic mapping, it currently suffers from inherent scale ambiguity due to its reliance on monocular visual inputs. Future work will incorporate multi-view extrinsic constraints and sensor fusion to achieve metric-accurate scale recovery, enabling robust navigation and interaction in complex robotic environments.
\bibliographystyle{IEEEtran}
\bibliography{references}

\newpage
\definecolor{best}{HTML}{FFC7CE}   
\definecolor{second}{HTML}{FFEB9C} 
\definecolor{third}{HTML}{FFFFCC}  
\twocolumn[
  \begin{center}
    {\huge \bf Supplementary Materials}
    \vspace{2em}
  \end{center}
]
\subsection{Camera Pose Estimation} 
\begin{figure*}[t]
    \centering
    \includegraphics[width=1.0\linewidth]{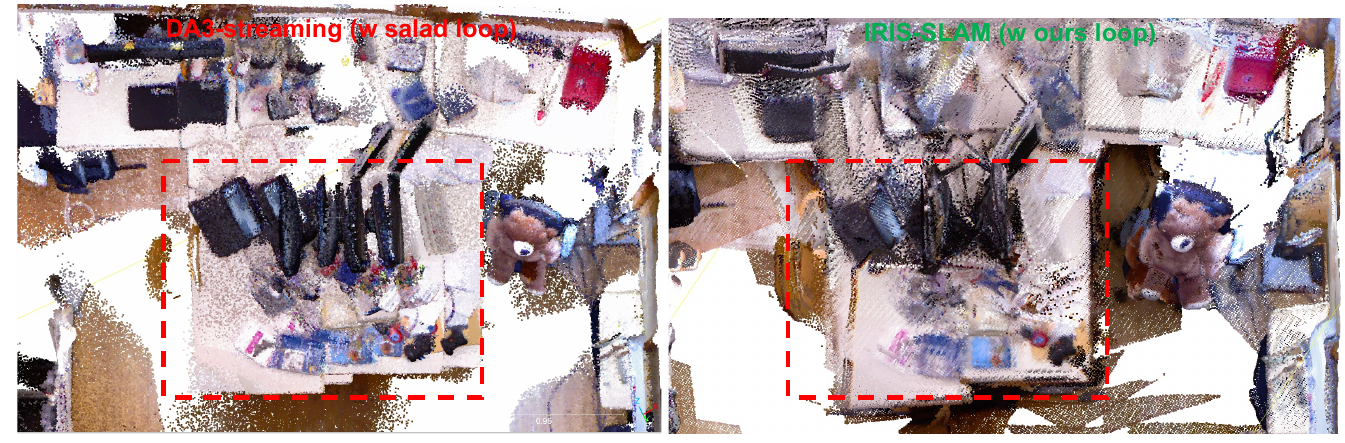}
    \caption{Qualitative comparison of reconstructed global point clouds on the TUM RGB-D fr1/room sequence: DepthAnything3 (DA3) Stream with Salad loop closure vs. our proposed IRIS-SLAM with instance-guided loop closure.}
    \label{fig:tum_room_vis}
\end{figure*}

\begin{figure*}[t]
    \centering
    \includegraphics[width=1.0\linewidth]{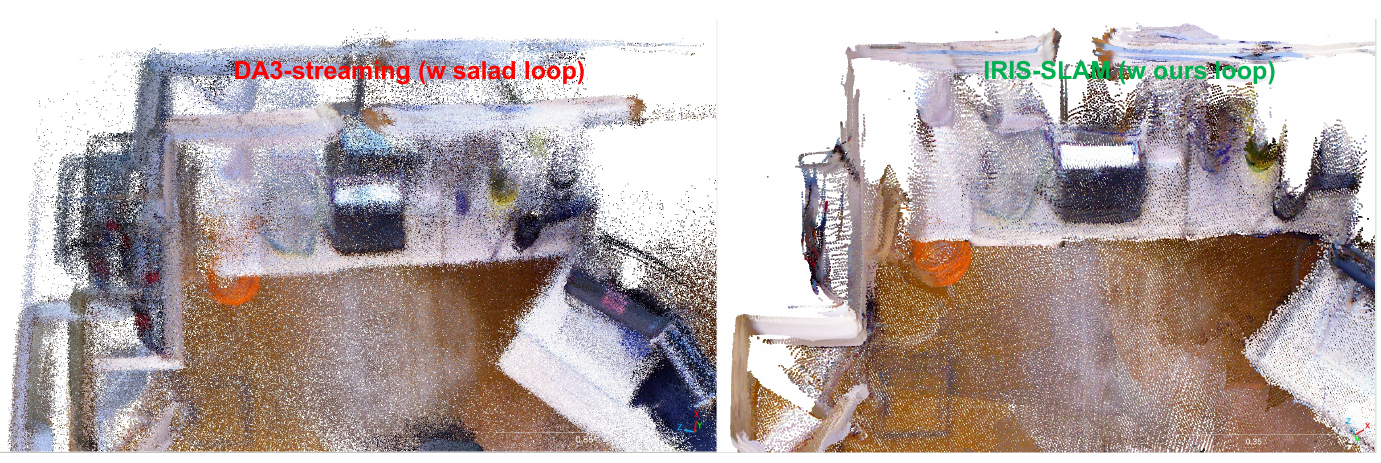}
    \caption{Qualitative comparison of reconstructed global point clouds on the TUM RGB-D fr1/teedy sequence.}
    \label{fig:tum_teddy_vis}
\end{figure*}

Beyond quantitative metrics, we conducted a qualitative comparison of geometric reconstruction quality against the state-of-the-art DA3-Stream (using DepthAnything3) on the fr1/room and fr1/teddy sequences, where our method's pose accuracy gains are most pronounced. As illustrated in Figures \ref{fig:tum_room_vis} and \ref{fig:tum_teddy_vis}, our instance-guided loop closure demonstrates superior robustness to viewpoint synthesis, enabling the construction of denser inter-chunk constraints. This effectively enhances the global consistency of the reconstructed map.

\subsection{3D Semantic Mapping}
Table \ref{tab:combined_eval} provides a detailed evaluation of semantic mapping performance on the ScanNet and Replica datasets. Our method leverages viewpoint-agnostic instance features as stable semantic anchors. These anchors allow the system to maintain robust association even when geometric co-visibility is negligible, ensuring superior global consistency in large-scale reconstructions. Furthermore, our approach represents a transition toward a unified SLAM architecture. Unlike baselines that require pre-computed ground-truth poses and depth, our framework operates as a cohesive, fully online streaming process.

\begin{figure*}[t] 
\centering 
\includegraphics[width=1.0\linewidth]{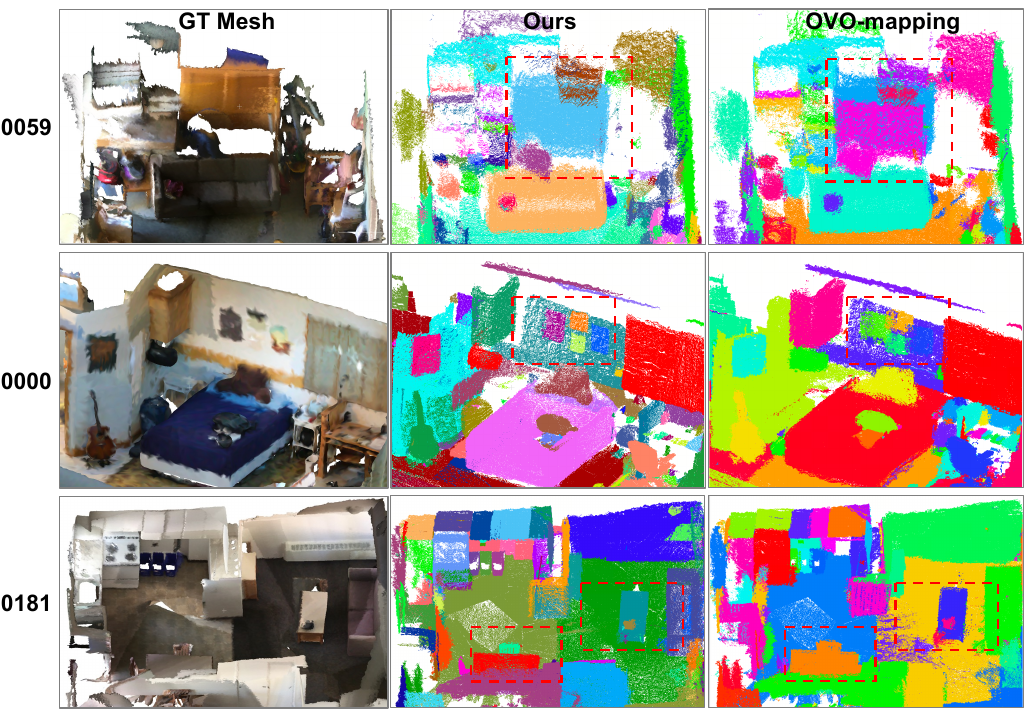} \caption{Qualitative comparison of semantic maps between IRIS-mapping (ours) and OVO-Mapping on ScanNet sequences; distinct colors represent unique object instance IDs. In scenes 0059 and 0181, OVO-Mapping fails to maintain stable instance associations, erroneously merging separate objects, such as foreground items with background cabinets or chairs with tables, into single color-coded entities. Conversely, our joint semantic-geometric tracking achieves precise instance separation. As demonstrated in scene 0000, our method also produces instance masks with significantly sharper boundaries and higher reconstruction fidelity than the baseline.} 
\label{fig:map_vis} 
\end{figure*}

As shown in Figure \ref{fig:map_vis}, we compare the semantic maps of IRIS-SLAM and OVO-Mapping on the ScanNet dataset. OVO-Mapping, the second-best performing method, frequently fails to distinguish spatially adjacent but semantically distinct objects. We see this in sequences 0059 and 0181. In sequence 0059, OVO-Mapping incorrectly merges a foreground object with a background cabinet. Similarly, in sequence 0181, it fails to separate a chair from a table. Our method effectively mitigates these issues through a joint semantic-geometric tracking module that maintains clear distinctions between different entities. Additionally, the results from sequence 0000 demonstrate that IRIS-SLAM captures much finer object contours, leading to more accurate and cleaner instance-level reconstructions.

\begin{table*}[t] 
\centering
\caption{Zero-shot semantic evaluations on ScanNet and Replica datasets. The best, second best, and third best results in each column are highlighted in \colorbox{best}{pink}, \colorbox{second}{orange}, and \colorbox{third}{yellow}, respectively. Our proposed methods are highlighted in \color{blue}{blue}.}
\label{tab:combined_eval}

\begin{minipage}{0.46\textwidth}
\centering
\subcaption{ScanNet dataset evaluation}
\resizebox{\linewidth}{!}{%
\setlength{\tabcolsep}{1.2pt} 
\begin{tabular}{lc cl ccccccc}
\toprule
\textbf{Method} & \textbf{Online} & \textbf{\shortstack{Pose/\\Depth}} & \textbf{Metric} & \textbf{0000} & \textbf{0059} & \textbf{0106} & \textbf{0169} & \textbf{0181} & \textbf{0207} & \textbf{Avg.} \\
\midrule
\multirow{4}{*}{\shortstack{Open-\\Gaussian}} & \multirow{4}{*}{\color{red}\xmark} & \multirow{4}{*}{GT/GT} & mIoU$\uparrow$ & 10.84 & 05.83 & 09.08 & 09.29 & 11.38 & 05.44 & 08.64 \\
& & & fIoU$\uparrow$ & 18.52 & 27.51 & 15.67 & 54.55 & 19.59 & 06.44 & 23.71 \\
& & & mAcc$\uparrow$ & 21.82 & 12.75 & 22.55 & 17.15 & 24.41 & 08.50 & 17.86 \\
& & & fAcc$\uparrow$ & 22.68 & 30.08 & 17.92 & 57.10 & 23.59 & 07.25 & 26.44 \\
\midrule
\multirow{4}{*}{\shortstack{Lang-\\Splat}} & \multirow{4}{*}{\color{red}\xmark} & \multirow{4}{*}{GT/GT} & mIoU$\uparrow$ & 06.72 & 07.41 & 08.55 & 06.02 & 11.13 & 03.48 & 07.22 \\
& & & fIoU$\uparrow$ & 20.03 & 46.98 & 29.48 & 43.60 & 23.06 & 02.38 & 27.59 \\
& & & mAcc$\uparrow$ & 18.99 & 22.48 & 25.14 & 25.39 & 28.87 & 05.18 & 21.01 \\
& & & fAcc$\uparrow$ & 26.26 & 52.80 & 36.09 & 47.34 & 28.21 & 02.54 & 32.21 \\
\midrule
\multirow{4}{*}{\shortstack{Grasp-\\Splats}} & \multirow{4}{*}{\color{green}\cmark} & \multirow{4}{*}{GT/GT} & mIoU$\uparrow$ & 06.27 & 08.22 & 06.46 & 08.40 & 05.53 & 05.27 & 06.69 \\
& & & fIoU$\uparrow$ & 08.39 & 25.88 & 19.26 & 30.77 & 07.58 & 07.82 & 16.62 \\
& & & mAcc$\uparrow$ & 12.27 & 20.24 & 09.46 & 26.28 & 10.80 & 10.29 & 14.89 \\
& & & fAcc$\uparrow$ & 09.60 & 28.87 & 21.12 & 38.66 & 09.79 & 09.40 & 19.57 \\
\midrule
\multirow{4}{*}{\shortstack{Omni-\\Map}} & \multirow{4}{*}{\color{green}\cmark} & \multirow{4}{*}{GT/GT} & mIoU$\uparrow$ & 20.25 & 14.81 & \cellcolor{third}38.44 & 24.41 & \cellcolor{third}33.04 & 21.59 & 25.42 \\
& & & fIoU$\uparrow$ & 34.24 & \cellcolor{third}53.59 & \cellcolor{best}64.18 & 57.34 & \cellcolor{second}54.96 & \cellcolor{second}40.86 & \cellcolor{second}50.86 \\
& & & mAcc$\uparrow$ & \cellcolor{second}49.81 & \cellcolor{third}47.36 & \cellcolor{second}62.37 & 41.43 & \cellcolor{best}61.61 & \cellcolor{second}42.99 & \cellcolor{second}50.93 \\
& & & fAcc$\uparrow$ & 40.95 & \cellcolor{third}61.26 & \cellcolor{third}71.72 & 62.77 & \cellcolor{second}59.90 & 45.68 & 57.05 \\
\midrule
\multirow{4}{*}{HOV-SG} & \multirow{4}{*}{\color{red}\xmark} & \multirow{4}{*}{GT/GT} & mIoU$\uparrow$ & 24.57 & 18.19 & 22.90 & 25.49 & 16.52 & 16.92 & 20.76 \\
& & & fIoU$\uparrow$ & 31.93 & 42.53 & 37.66 & \cellcolor{second}58.84 & 28.58 & 30.51 & 38.34 \\
& & & mAcc$\uparrow$ & 46.78 & 37.98 & 47.41 & \cellcolor{third}45.21 & 37.08 & 34.56 & 41.50 \\
& & & fAcc$\uparrow$ & 38.74 & 51.73 & 42.41 & \cellcolor{third}69.97 & 33.47 & 36.65 & 45.50 \\
\midrule
\multirow{4}{*}{\shortstack{Concept-\\Fusion}} & \multirow{4}{*}{\color{green}\cmark} & \multirow{4}{*}{GT/GT} & mIoU$\uparrow$ & 11.13 & 05.48 & 09.55 & 06.15 & 09.92 & 08.77 & 08.50 \\
& & & fIoU$\uparrow$ & 23.70 & 32.18 & 37.65 & 42.38 & 29.15 & 15.24 & 30.05 \\
& & & mAcc$\uparrow$ & 34.92 & 28.69 & 37.16 & 30.35 & 35.65 & 24.06 & 31.81 \\
& & & fAcc$\uparrow$ & 31.50 & 37.36 & 45.10 & 47.92 & 36.52 & 21.41 & 36.64 \\
\midrule
\multirow{4}{*}{\shortstack{Concept-\\Graph}} & \multirow{4}{*}{\color{green}\cmark} & \multirow{4}{*}{GT/GT} & mIoU$\uparrow$ & 15.38 & 16.06 & 18.57 & 17.78 & 16.55 & 13.42 & 16.29 \\
& & & fIoU$\uparrow$ & 16.17 & 42.06 & 39.11 & 52.52 & 30.56 & 19.31 & 33.29 \\
& & & mAcc$\uparrow$ & 35.69 & 32.87 & 40.52 & 37.67 & 32.09 & 25.60 & 34.07 \\
& & & fAcc$\uparrow$ & 21.84 & 53.36 & 47.93 & 64.08 & 38.27 & 24.15 & 41.60 \\
\midrule
\multirow{4}{*}{\shortstack{Open-\\Fusion}} & \multirow{4}{*}{\color{green}\cmark} & \multirow{4}{*}{GT/GT} & mIoU$\uparrow$ & 18.58 & 12.09 & 15.76 & 19.32 & 21.26 & 21.11 & 18.02 \\
& & & fIoU$\uparrow$ & 28.56 & 48.04 & 36.22 & \cellcolor{best}59.03 & \cellcolor{third}50.44 & \cellcolor{third}40.60 & \cellcolor{third}43.82 \\
& & & mAcc$\uparrow$ & \cellcolor{third}47.85 & \cellcolor{second}44.21 & 42.95 & 39.06 & \cellcolor{third}43.27 & \cellcolor{best}48.52 & 44.31 \\
& & & fAcc$\uparrow$ & 38.84 & 56.89 & 39.51 & 62.63 & \cellcolor{third}57.10 & \cellcolor{second}51.56 & 51.09 \\
\midrule
\multirow{4}{*}{\shortstack{OVO-\\mapping}} & \multirow{4}{*}{\color{green}\cmark} & \multirow{4}{*}{GT/GT} & mIoU$\uparrow$ & \cellcolor{second}29.61 & \cellcolor{second}39.40 & \cellcolor{second}43.65 & 28.74 & 24.19 & \cellcolor{best}23.89 & \cellcolor{third}31.58 \\
& & & fIoU$\uparrow$ & \cellcolor{second}48.27 & \cellcolor{second}59.00 & 54.26 & 38.74 & 39.99 & \cellcolor{best}44.30 & \cellcolor{third}47.43 \\
& & & mAcc$\uparrow$ & 44.47 & \cellcolor{best}52.78 & \cellcolor{third}58.10 & 40.40 & 36.66 & \cellcolor{third}40.85 & \cellcolor{third}45.54 \\
& & & fAcc$\uparrow$ & \cellcolor{second}62.30 & \cellcolor{second}72.42 & \cellcolor{third}68.38 & 56.23 & 52.84 & \cellcolor{best}57.28 & \cellcolor{second}61.58 \\
\midrule
\multirow{4}{*}{\shortstack{\textbf{\color{blue}IRIS-}\\\textbf{\color{blue}mapping}}} & \multirow{4}{*}{\color{green}\cmark} & \multirow{4}{*}{GT/GT} & mIoU$\uparrow$ & \cellcolor{best}36.66 & \cellcolor{best}50.55 & \cellcolor{best}50.81 & \cellcolor{second}36.51 & \cellcolor{best}41.26 & \cellcolor{second}23.80 & \cellcolor{best}39.93 \\
& & & fIoU$\uparrow$ & \cellcolor{best}54.23 & \cellcolor{best}63.51 & \cellcolor{second}61.92 & 45.45 & \cellcolor{best}59.72 & 36.87 & \cellcolor{best}53.62 \\
& & & mAcc$\uparrow$ & \cellcolor{best}57.05 & \cellcolor{best}68.35 & \cellcolor{best}62.74 & \cellcolor{second}49.26 & \cellcolor{second}53.54 & \cellcolor{second}41.24 & \cellcolor{best}55.36 \\
& & & fAcc$\uparrow$ & \cellcolor{best}68.40 & \cellcolor{best}75.68 & \cellcolor{best}73.86 & 63.20 & \cellcolor{best}69.89 & 49.63 & \cellcolor{best}66.78 \\
\midrule
\multirow{4}{*}{\shortstack{\textbf{\color{blue}IRIS-}\\\textbf{\color{blue}SLAM}}} & \multirow{4}{*}{\color{green}\cmark} & \multirow{4}{*}{\textbf{\color{blue}Est/Est}} & mIoU$\uparrow$ & \cellcolor{third}29.24 & \cellcolor{third}25.58 & \cellcolor{third}43.02 & \cellcolor{best}43.91 & \cellcolor{second}25.92 & \cellcolor{third}22.02 & \cellcolor{second}31.62 \\
& & & fIoU$\uparrow$ & \cellcolor{third}44.49 & 27.91 & \cellcolor{third}57.92 & \cellcolor{third}58.04 & \cellcolor{third}40.70 & 34.05 & 43.85 \\
& & & mAcc$\uparrow$ & \cellcolor{third}45.61 & 38.14 & \cellcolor{third}55.30 & \cellcolor{best}54.19 & 37.08 & 40.14 & \cellcolor{third}45.08 \\
& & & fAcc$\uparrow$ & \cellcolor{third}60.11 & 46.58 & \cellcolor{second}71.76 & \cellcolor{best}74.18 & \cellcolor{third}51.50 & 45.81 & \cellcolor{third}58.32 \\
\bottomrule
\end{tabular}%
}
\end{minipage}
\hfill 
\begin{minipage}{0.53\textwidth}
\centering
\subcaption{Replica dataset evaluation}
\resizebox{\linewidth}{!}{
\setlength{\tabcolsep}{1.2pt} 
\begin{tabular}{l c l l c c c c c c c c c}
\toprule

\textbf{Method} & \textbf{Online} & \makecell[c]{\textbf{Pose}\\\textbf{/Depth}} & \textbf{Metric} & \textbf{ro-0} & \textbf{ro-1} & \textbf{ro-2} & \textbf{of-0} & \textbf{of-1} & \textbf{of-2} & \textbf{of-3} & \textbf{of-4} & \textbf{Avg.} \\
\midrule
\multirow{4}{*}{\makecell[l]{Open-\\Gaussian}} & \multirow{4}{*}{\color{red}\xmark} & \multirow{4}{*}{\makecell[c]{GT/GT}} & mIoU$\uparrow$ & 11.64 & 03.82 & 09.04 & 06.06 & 04.97 & 02.10 & 05.55 & 11.36 & 06.82 \\
& & & fIoU$\uparrow$ & 31.07 & 10.44 & 22.17 & 10.14 & 15.32 & 03.15 & 07.00 & 24.00 & 15.41 \\
& & & mAcc$\uparrow$ & 29.39 & 11.13 & 19.67 & 10.07 & 20.36 & 07.01 & 11.45 & 24.16 & 16.66 \\
& & & fAcc$\uparrow$ & 05.96 & 14.35 & 28.09 & 12.24 & 16.90 & 04.92 & 07.90 & 24.31 & 18.08 \\
\midrule
\multirow{4}{*}{\makecell[l]{Lang-\\Splat}} & \multirow{4}{*}{\color{red}\xmark} & \multirow{4}{*}{\makecell[c]{GT/GT}} & mIoU$\uparrow$ & 12.62 & 11.30 & 10.48 & 10.40 & 05.38 & 10.33 & 09.55 & 09.91 & 10.00 \\
& & & fIoU$\uparrow$ & 47.83 & 47.74 & 41.56 & 29.69 & 24.49 & 54.13 & 44.79 & 27.29 & 39.69 \\
& & & mAcc$\uparrow$ & 26.76 & 28.71 & 25.09 & 21.85 & 14.73 & 22.71 & 19.90 & 23.68 & 22.93 \\
& & & fAcc$\uparrow$ & 53.77 & 53.70 & 50.51 & 31.94 & 28.57 & 56.78 & 50.24 & 27.77 & 44.16 \\
\midrule
\multirow{4}{*}{\makecell[l]{Grasp-\\Splats}} & \multirow{4}{*}{\color{green}\cmark} & \multirow{4}{*}{\makecell[c]{GT/GT}} & mIoU$\uparrow$ & 16.74 & 08.25 & 14.13 & 11.35 & 02.77 & 11.13 & 14.79 & 04.17 & 10.42 \\
& & & fIoU$\uparrow$ & 55.05 & 30.75 & 48.56 & 51.38 & 14.12 & 45.22 & 45.37 & 50.88 & 42.67 \\
& & & mAcc$\uparrow$ & 32.92 & 21.74 & 32.45 & 33.51 & 08.89 & 22.12 & 27.92 & 10.80 & 23.79 \\
& & & fAcc$\uparrow$ & 66.82 & 45.24 & 65.38 & 59.56 & 22.42 & 52.92 & 54.60 & 52.13 & 52.39 \\
\midrule
\multirow{4}{*}{\makecell[l]{Omni-\\Map}} & \multirow{4}{*}{\color{green}\cmark} & \multirow{4}{*}{\makecell[c]{GT/GT}} & mIoU$\uparrow$ & \cellcolor{best}49.57 & \cellcolor{best}42.63 & \cellcolor{third}26.50 & \cellcolor{third}21.92 & \cellcolor{second}21.32 & \cellcolor{third}24.45 & \cellcolor{third}23.31 & \cellcolor{third}22.82 & \cellcolor{third}29.06 \\
& & & fIoU$\uparrow$ & \cellcolor{second}70.73 & \cellcolor{third}57.89 & \cellcolor{third}65.56 & \cellcolor{best}62.37 & \cellcolor{best}52.14 & \cellcolor{best}76.53 & \cellcolor{second}62.30 & \cellcolor{third}67.85 & \cellcolor{best}64.42 \\
& & & mAcc$\uparrow$ & \cellcolor{second}62.94 & \cellcolor{best}61.68 & \cellcolor{second}41.04 & \cellcolor{best}38.28 & \cellcolor{third}32.76 & \cellcolor{best}39.42 & \cellcolor{best}37.17 & \cellcolor{third}43.03 & \cellcolor{best}44.54 \\
& & & fAcc$\uparrow$ & \cellcolor{best}82.29 & \cellcolor{third}69.93 & \cellcolor{third}74.66 & \cellcolor{second}71.40 & \cellcolor{second}55.29 & \cellcolor{best}81.94 & \cellcolor{third}70.85 & \cellcolor{third}71.37 & \cellcolor{second}72.22 \\
\midrule
\multirow{4}{*}{HOV-SG} & \multirow{4}{*}{\color{red}\xmark} & \multirow{4}{*}{\makecell[c]{GT/GT}} & mIoU$\uparrow$ & \cellcolor{second}36.26 & 29.00 & 24.13 & 19.14 & \cellcolor{best}22.69 & 23.88 & 15.28 & 19.89 & 23.79 \\
& & & fIoU$\uparrow$ & \cellcolor{best}71.24 & 51.95 & 62.52 & 44.05 & \cellcolor{second}49.50 & 46.28 & 20.95 & 44.42 & 48.86 \\
& & & mAcc$\uparrow$ & \cellcolor{third}49.68 & \cellcolor{third}52.96 & \cellcolor{third}39.79 & 30.60 & \cellcolor{best}40.28 & \cellcolor{second}35.16 & \cellcolor{third}28.86 & 39.36 & \cellcolor{third}39.59 \\
& & & fAcc$\uparrow$ & \cellcolor{second}81.69 & 61.36 & 72.31 & 46.72 & \cellcolor{best}59.72 & 47.81 & 24.90 & 46.71 & 55.15 \\
\midrule
\multirow{4}{*}{\makecell[l]{Concept-\\Fusion}} & \multirow{4}{*}{\color{green}\cmark} & \multirow{4}{*}{\makecell[c]{GT/GT}} & mIoU$\uparrow$ & 07.96 & 08.86 & 02.57 & 04.66 & 03.82 & 02.93 & 03.53 & 03.68 & 04.75 \\
& & & fIoU$\uparrow$ & 32.28 & 34.86 & 31.65 & 23.45 & 20.35 & 23.35 & 19.15 & 17.29 & 25.30 \\
& & & mAcc$\uparrow$ & 25.41 & 31.04 & 07.59 & 17.42 & 23.38 & 12.13 & 16.65 & 20.71 & 19.29 \\
& & & fAcc$\uparrow$ & 38.11 & 42.19 & 37.44 & 24.86 & 22.82 & 24.43 & 21.85 & 20.23 & 28.99 \\
\midrule
\multirow{4}{*}{\makecell[l]{Concept-\\Graph}} & \multirow{4}{*}{\color{green}\cmark} & \multirow{4}{*}{\makecell[c]{GT/GT}} & mIoU$\uparrow$ & 22.53 & 18.74 & 14.66 & \cellcolor{third}19.36 & 11.22 & 15.75 & 11.66 & 17.74 & 16.46 \\
& & & fIoU$\uparrow$ & 50.28 & 46.21 & 54.52 & 43.87 & 22.00 & 23.25 & 11.32 & 34.07 & 35.69 \\
& & & mAcc$\uparrow$ & 38.88 & 36.57 & 25.10 & 29.32 & 22.54 & \cellcolor{third}33.65 & 28.43 & 37.56 & 31.51 \\
& & & fAcc$\uparrow$ & 59.65 & 57.93 & 62.64 & 50.21 & 25.64 & 27.78 & 16.05 & 39.63 & 42.44 \\
\midrule
\multirow{4}{*}{\makecell[l]{Open-\\Fusion}} & \multirow{4}{*}{\color{green}\cmark} & \multirow{4}{*}{\makecell[c]{GT/GT}} & mIoU$\uparrow$ & 22.17 & 16.64 & 21.94 & 08.95 & 12.78 & 12.36 & 18.10 & 18.00 & 16.37 \\
& & & fIoU$\uparrow$ & 60.17 & 41.62 & 61.72 & \cellcolor{third}49.64 & \cellcolor{third}33.86 & 48.22 & 53.00 & 64.93 & 51.65 \\
& & & mAcc$\uparrow$ & \cellcolor{third}41.65 & \cellcolor{third}41.39 & \cellcolor{best}43.03 & 25.61 & 30.01 & 26.99 & \cellcolor{third}32.14 & 40.40 & 35.15 \\
& & & fAcc$\uparrow$ & 72.78 & 54.68 & \cellcolor{third}73.71 & \cellcolor{third}60.10 & \cellcolor{third}37.01 & \cellcolor{third}52.82 & \cellcolor{third}62.53 & 69.33 & 60.37 \\
\midrule
\multirow{4}{*}{\makecell[l]{OVO-\\mapping}} & \multirow{4}{*}{\color{green}\cmark} & \multirow{4}{*}{\makecell[c]{GT/GT}} & mIoU$\uparrow$ & \cellcolor{third}31.08 & \cellcolor{third}37.49 & \cellcolor{second}29.33 & \cellcolor{second}23.60 & 16.36 & \cellcolor{best}29.88 & \cellcolor{best}29.20 & \cellcolor{second}42.50 & \cellcolor{second}29.93 \\
& & & fIoU$\uparrow$ & 40.84 & \cellcolor{best}67.48 & \cellcolor{best}68.66 & 47.23 & 16.52 & \cellcolor{second}69.76 & \cellcolor{best}65.24 & \cellcolor{second}80.30 & \cellcolor{third}57.00 \\
& & & mAcc$\uparrow$ & \cellcolor{best}62.46 & \cellcolor{second}53.74 & 37.49 & \cellcolor{third}33.08 & \cellcolor{second}33.11 & \cellcolor{second}36.18 & \cellcolor{second}35.32 & \cellcolor{second}57.43 & \cellcolor{second}43.60 \\
& & & fAcc$\uparrow$ & \cellcolor{third}74.98 & \cellcolor{second}78.77 & \cellcolor{second}80.00 & 59.65 & 32.34 & 32.34 & \cellcolor{best}77.26 & \cellcolor{second}87.78 & 65.39 \\
\midrule
\multirow{4}{*}{\makecell[l]{\textbf{\color{blue}IRIS-}\\\textbf{\color{blue}mapping}}} & \multirow{4}{*}{\color{green}\cmark} & \multirow{4}{*}{\makecell[c]{GT/GT}} & mIoU$\uparrow$ & 28.43 & \cellcolor{second}40.25 & \cellcolor{best}30.25 & \cellcolor{best}25.78 & \cellcolor{third}20.46 & \cellcolor{second}28.69 & \cellcolor{second}28.40 & \cellcolor{best}46.96 & \cellcolor{best}31.15 \\
& & & fIoU$\uparrow$ & \cellcolor{third}61.41 & \cellcolor{second}67.54 & \cellcolor{second}68.58 & \cellcolor{second}48.72 & 20.62 & \cellcolor{third}68.98 & \cellcolor{third}62.50 & \cellcolor{best}81.67 & \cellcolor{second}60.00 \\
& & & mAcc$\uparrow$ & 36.09 & \cellcolor{third}55.65 & \cellcolor{third}37.63 & \cellcolor{second}35.31 & 31.43 & \cellcolor{third}34.23 & \cellcolor{second}36.81 & \cellcolor{best}57.92 & \cellcolor{third}40.63 \\
& & & fAcc$\uparrow$ & \cellcolor{third}75.34 & \cellcolor{best}79.65 & \cellcolor{best}81.05 & \cellcolor{best}60.76 & 35.83 & \cellcolor{second}81.61 & \cellcolor{second}75.48 & \cellcolor{best}88.65 & \cellcolor{best}72.30 \\
\midrule
\multirow{4}{*}{\makecell[l]{\textbf{\color{blue}IRIS-}\\\textbf{\color{blue}SLAM}}} & \multirow{4}{*}{\color{green}\cmark} & \multirow{4}{*}{\makecell[c]{\textbf{\color{blue}Est/Est}}} & mIoU$\uparrow$ & 22.16 & 21.77 & 20.28 & 16.92 & 10.61 & \cellcolor{third}23.94 & 16.06 & \cellcolor{third}30.03 & 20.22 \\
& & & fIoU$\uparrow$ & 50.41 & 45.59 & 52.41 & 42.46 & 23.06 & \cellcolor{third}62.46 & \cellcolor{third}56.58 & \cellcolor{third}69.24 & 50.28 \\
& & & mAcc$\uparrow$ & 31.00 & 34.24 & 27.40 & 25.43 & 19.77 & \cellcolor{third}33.11 & 23.84 & \cellcolor{third}40.68 & 29.43 \\
& & & fAcc$\uparrow$ & 66.53 & \cellcolor{third}62.37 & 69.85 & 57.60 & \cellcolor{third}39.01 & \cellcolor{third}75.78 & \cellcolor{third}69.71 & \cellcolor{third}79.13 & \cellcolor{third}65.00 \\\\\bottomrule
\end{tabular}
}
\end{minipage}

\end{table*}


\begin{table*}[t] 
\centering

\begin{minipage}[t]{0.58\linewidth}
    \centering
    \caption{Computational efficiency on ScanNet (chunk size = 120 frames). The average per-frame processing time is approximately 73 ms, achieving a real-time throughput of 13.9 FPS.}
    \label{tab:runtime_analysis}
    \setlength{\tabcolsep}{4pt} 
    \begin{tabular}{lcccccc}
    \toprule
    Unit (ms) & Inference & \begin{tabular}[c]{@{}c@{}}Mask\\ Seg.\end{tabular} & \begin{tabular}[c]{@{}c@{}}Inst.\\ Assoc.\end{tabular} & \begin{tabular}[c]{@{}c@{}}Chunk\\ Align.\end{tabular} & \begin{tabular}[c]{@{}c@{}}Loop\\ Det.\end{tabular} & \begin{tabular}[c]{@{}c@{}}Loop\\ Opt.\end{tabular} \\ \midrule
    Per-chunk & 152.50 & 130.08 & 5132.39 & 1798.30 & 780.13 & 766.83 \\
    Per-frame & 1.27 & 1.09 & 42.77 & 14.99 & 6.50 & 6.39 \\ \bottomrule
    \end{tabular}
\end{minipage}
\hfill 
\begin{minipage}[t]{0.38\linewidth}
    \centering
    \caption{Comparison of clustering execution time between our method and IGGT. All experiments were conducted on an NVIDIA RTX 4090. 'OOM' denotes Out-of-Memory.}
    \label{tab:clustering_time}
    \setlength{\tabcolsep}{8pt}
    \begin{tabular}{l c c}
        \toprule
        \multirow{2}{*}{Method} & \multicolumn{2}{c}{Clustering Time (s)} \\
        \cmidrule(lr){2-3} 
         & 8 frames & 120 frames \\
        \midrule
        IGGT & 77.642 & OOM \\
        Ours & \textbf{0.0244} & \textbf{0.1308} \\
        \bottomrule
    \end{tabular}
\end{minipage}

\end{table*}

\subsection{Instance-guided Loop Closure}
\subsubsection{Ground Truth Criteria for Loop Closure}
To establish a rigorous evaluation foundation, we define true positive loop closures through strict geometric co-visibility verification.
Following the methodology of SLC²-SLAM \cite{10935649}, we first construct a loop closure evaluation benchmark comprising query and place frames. To determine the true positive loop closures, we adopt the overlap rate from OverlapNet \cite{chen2020rss} as our primary metric. This metric leverages camera intrinsics and poses to back-project observation points from the source frame into 3D space, which are then re-projected onto the target frame's imaging plane. By performing visibility and depth consistency checks, we quantify the effective overlapping area. This geometry-based ground truth generation accounts for both scene consistency and viewpoint variations. It enables the robust identification of loop pairs even under significant viewpoint shifts, provided partial overlap remains, thereby ensuring the reliability of the evaluation in wide-baseline scenarios.

\begin{figure*}[t]
    \centering
    \begin{minipage}[b]{0.49\linewidth}
        \centering
        \includegraphics[width=\linewidth]{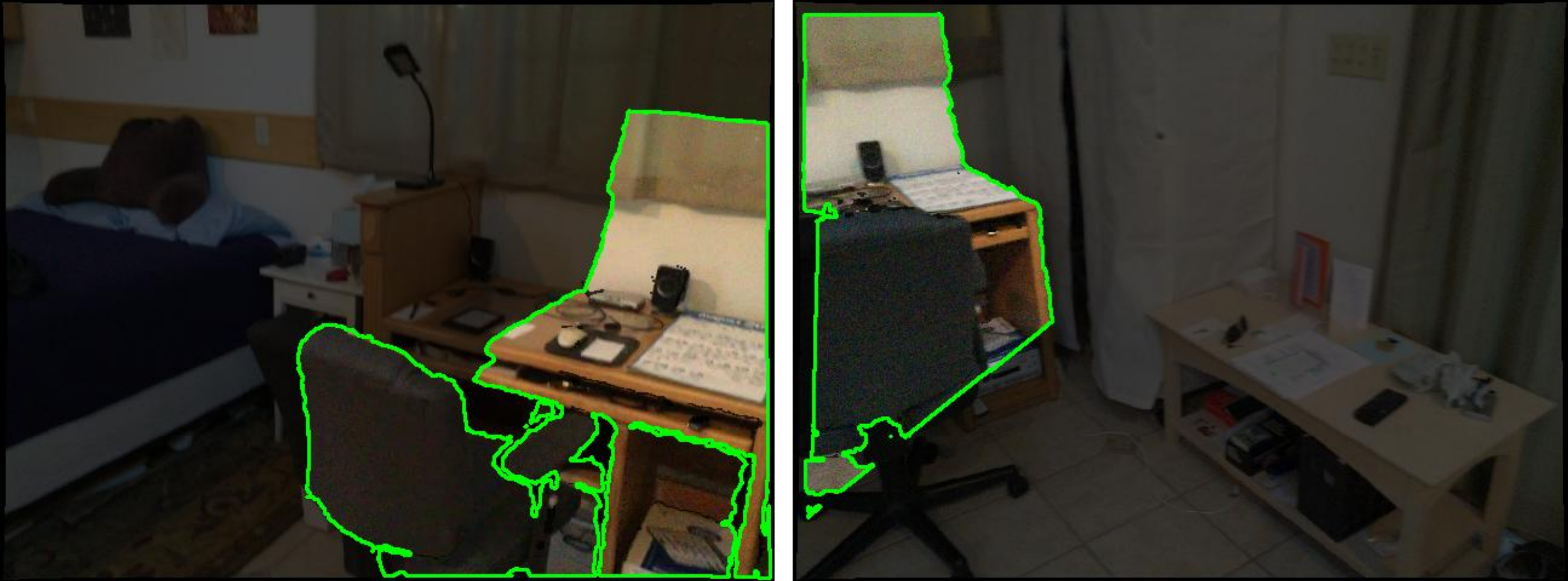}
    \end{minipage}
    \hfill
    \begin{minipage}[b]{0.49\linewidth}
        \centering
        \includegraphics[width=\linewidth]{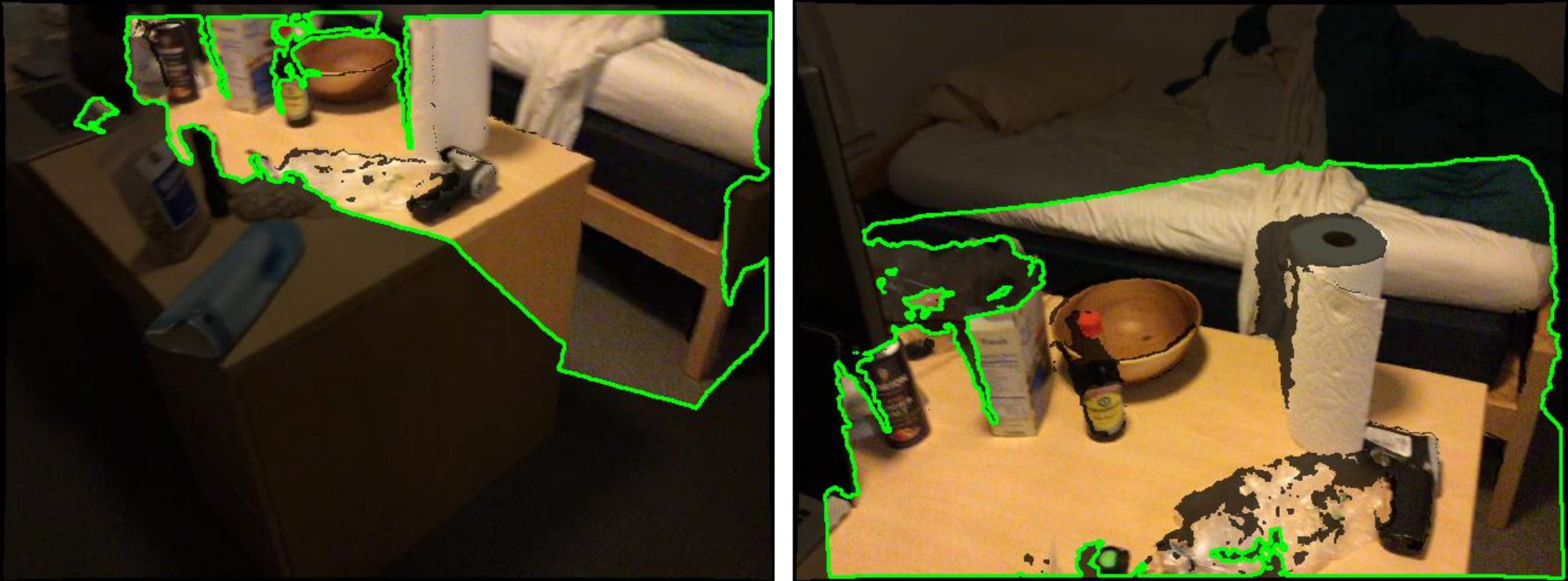}
    \end{minipage}

    \caption{Visualization of the overlap area calculated via our proposed scheme. The figure presents two sets of concatenated pairs consisting of a query frame and a place frame. The green boundaries delineate the comprehensive co-visible regions identified by our method, with non-overlapping areas dimmed for clarity.}
    \label{fig:overlap_visualization}
\end{figure*}

A pair of images is classified as a true positive loop closure when their overlap exceeds a predefined threshold. As shown in Fig. \ref{fig:overlap_visualization}, both image pairs constitute valid loop closures based on the overlap metric.

\subsubsection{Quantitative and Qualitative Evaluation}

\begin{figure*}[t]
    \centering
    \begin{minipage}[b]{0.48\textwidth}
        \centering
        \includegraphics[width=\linewidth]{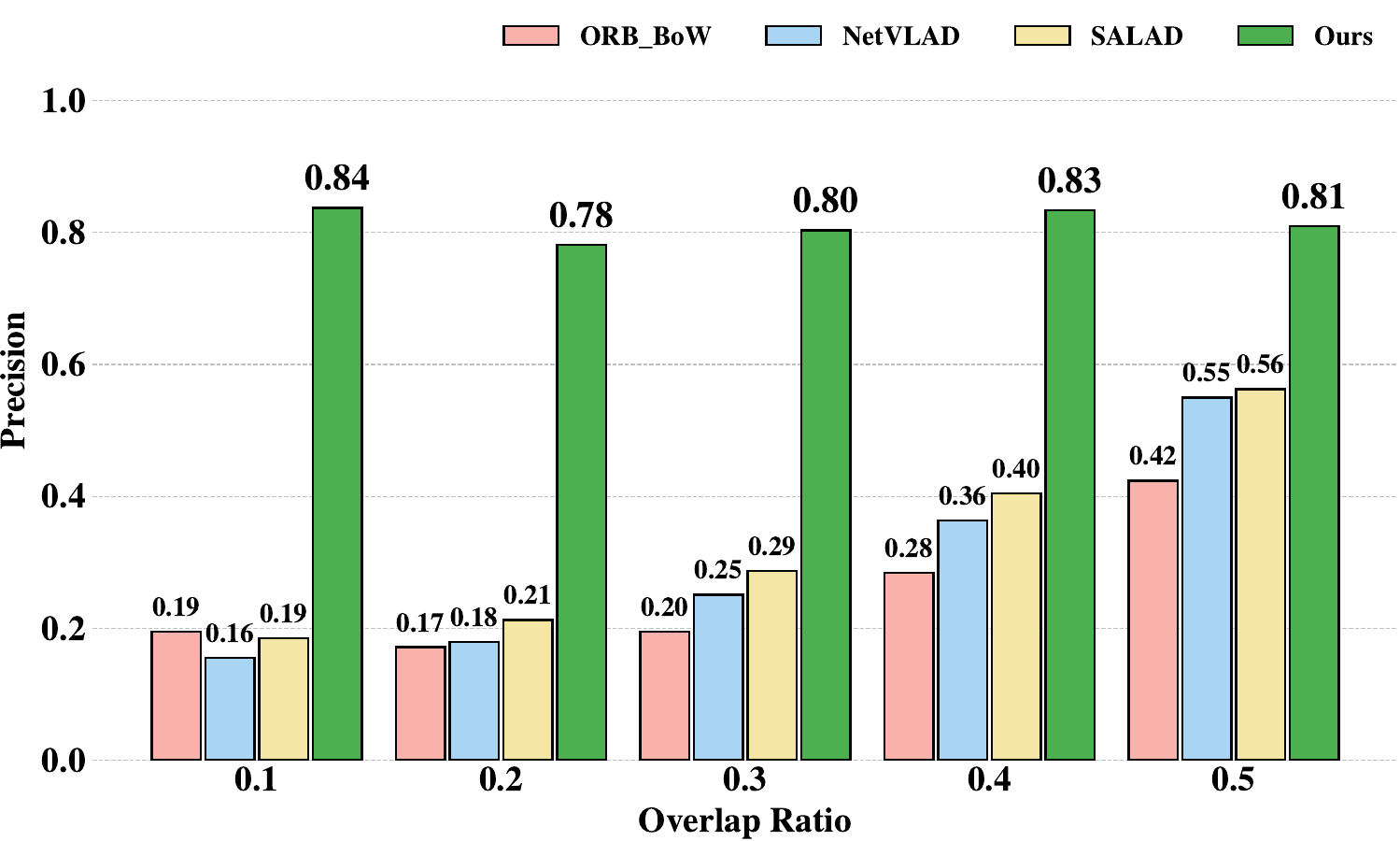}
    \end{minipage}
    \hfill
    \begin{minipage}[b]{0.48\textwidth}
        \centering
        \includegraphics[width=\linewidth]{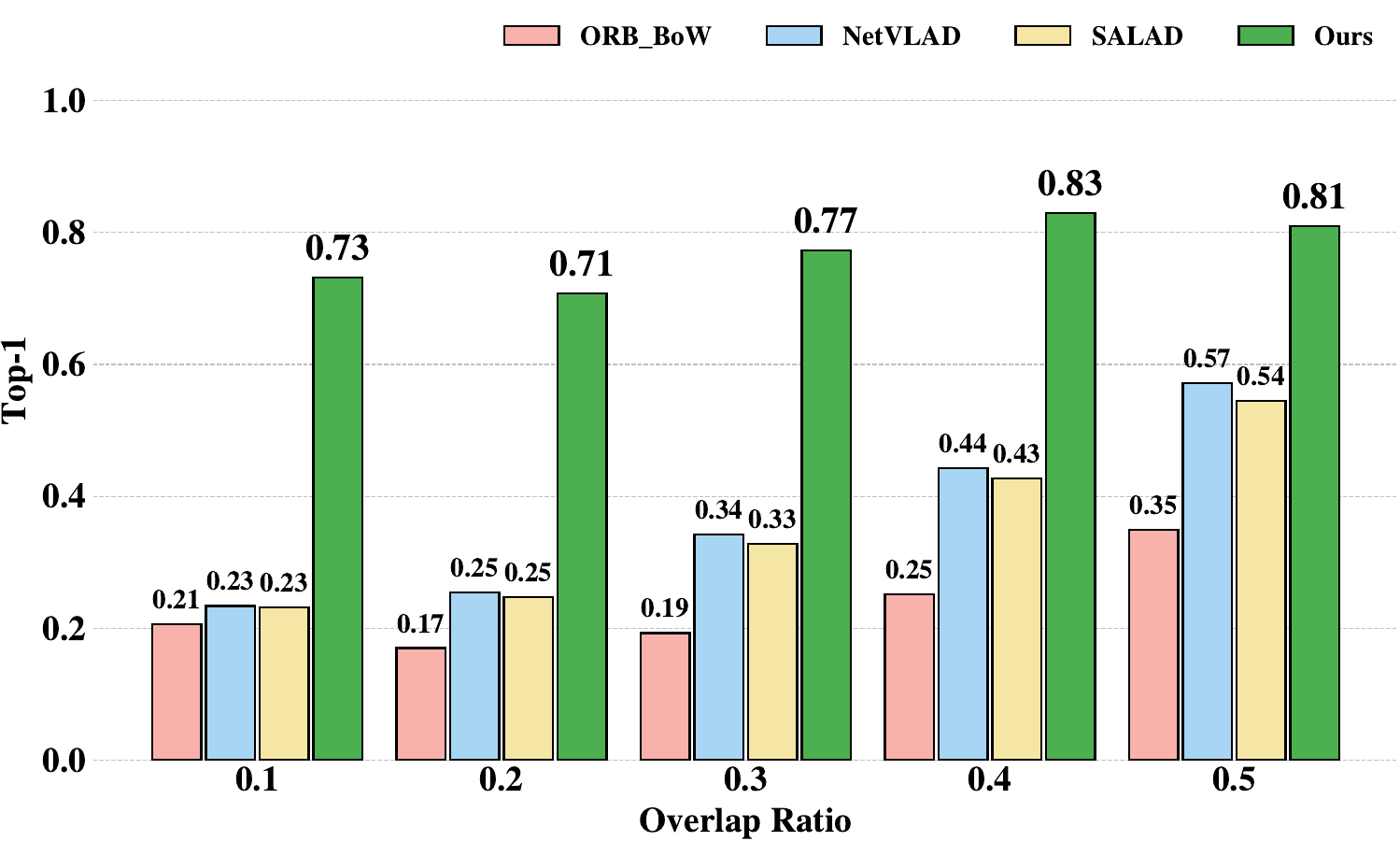}
    \end{minipage}
    \caption{Comparison of average Precision and Top-1 metrics across overlap ratio gradients from $0.1$ to $0.5$. The results, averaged across selected ScanNet sequences, demonstrate the significant robustness of our method in wide-baseline scenarios with low visual overlap.}
    \label{fig:supplementary_metrics}
\end{figure*}

Figure \ref{fig:supplementary_metrics} further compares the Precision and Top-1 retrieval performance. IRIS-SLAM demonstrates a substantial improvement in both metrics compared to existing baselines. Notably, in low-overlap regimes where visual features are sparse, the precision of baseline methods often falls below 0.2, while our approach maintains a consistently higher precision of approximately 0.8. 

\begin{figure*}[t]
    \centering
    \includegraphics[width=\textwidth]{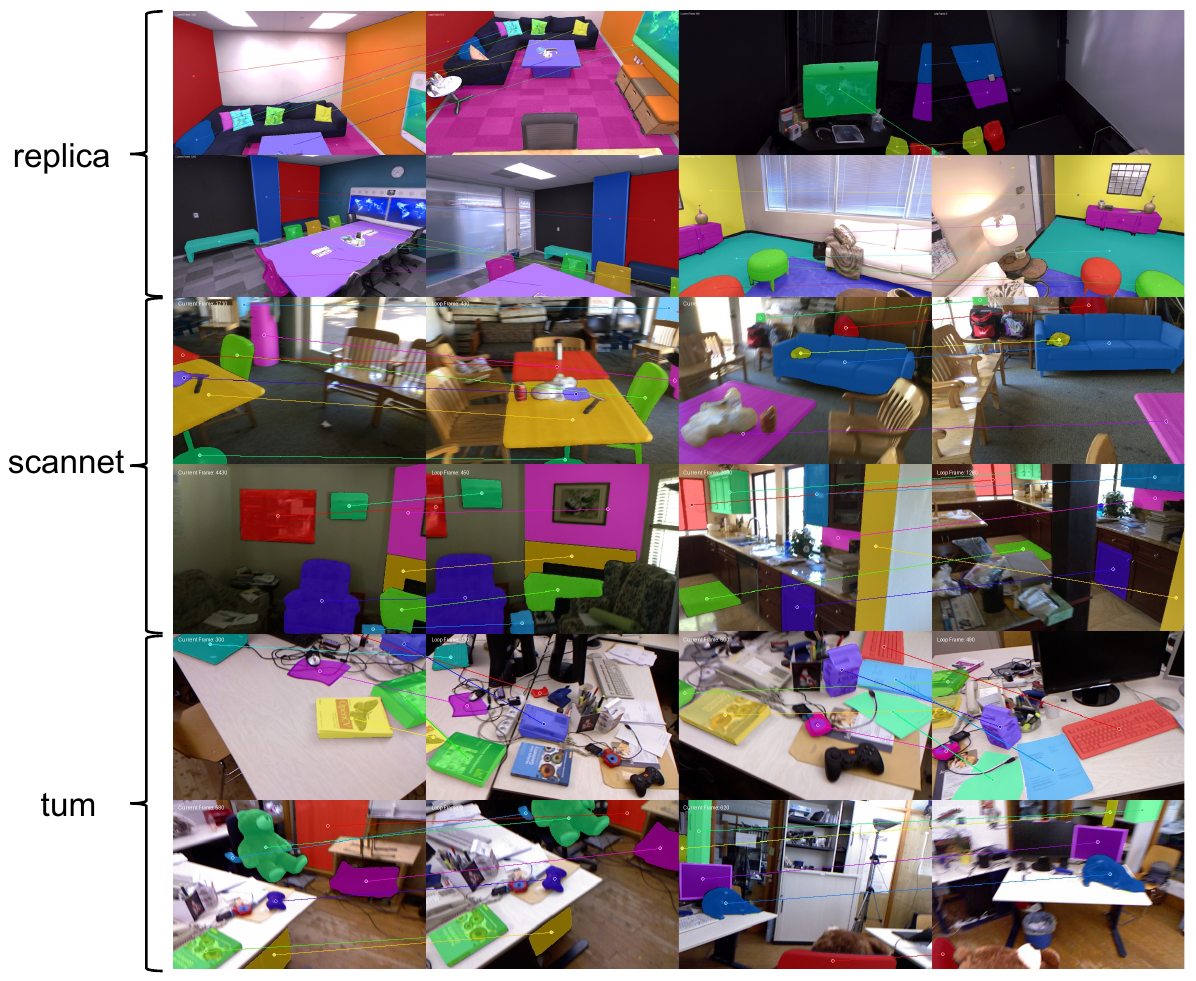} 
    \caption{Robust loop closure detection under challenging large-viewpoint variations on Replica (top), ScanNet (middle), and TUM (bottom) datasets. By leveraging viewpoint-agnostic semantic anchors, our approach achieves significantly higher loop closure reliability than appearance-based methods under large viewpoint changes.}
    \label{fig:loop_matching}
\end{figure*}

Figure \ref{fig:loop_matching} further demonstrates the robust loop closure detection capability of IRIS-SLAM across large viewpoint change on the Replica, ScanNet, and TUM datasets, where traditional methods fail to detect these loop closures.

\begin{figure*}[t]
    \centering
    \includegraphics[width=\textwidth]{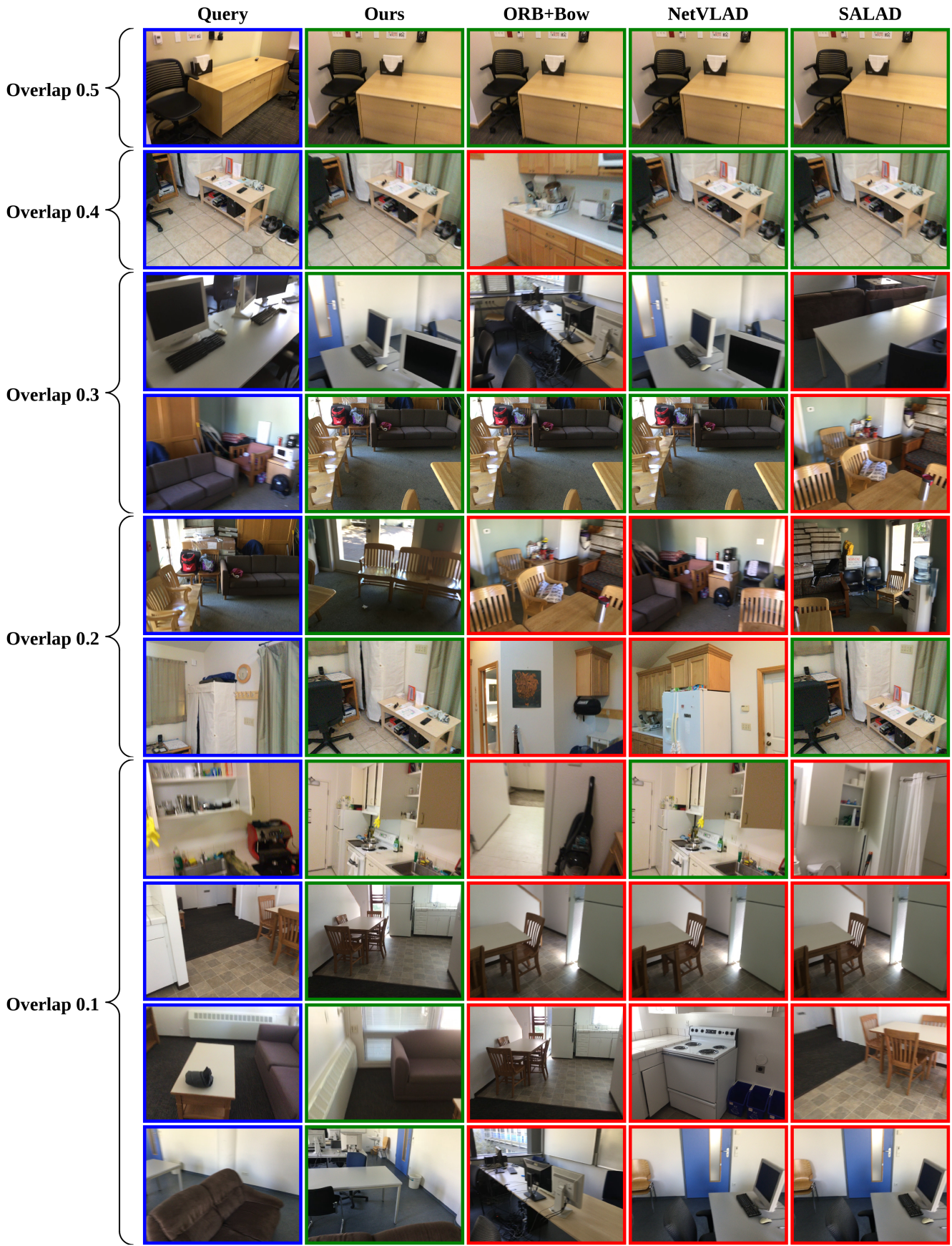}  
    \caption{Comparison of loop closure detection results under different overlaps. As the overlap decreases, conventional methods increasingly return incorrect loop closures, whereas our Instance-guided method is able to return correct loop closures across the tested scenarios.}
    \label{fig:qualitative_comparison}
\end{figure*}

The qualitative comparisons on our evaluation benchmark are illustrated in Figure \ref{fig:qualitative_comparison}, where green and red bounding boxes denote correct and erroneous loop closures, respectively. The visualization and comparisons demonstrate that leveraging semantic instance matching can mitigate the impact of viewpoint changes even in low-overlap scenarios.

\subsection{Computational Efficiency Analysis of IRIS-SLAM}
We evaluated the computational efficiency of IRIS-SLAM on six ScanNet sequences using a 120-frame chunk size. Table \ref{tab:runtime_analysis} shows that our system maintains exceptionally low per-frame processing times; core tasks like inference and mask segmentation require only 1.27 ms and 1.09 ms, respectively. While chunk-level operations (e.g., instance association and alignment) are more complex, they only add a manageable per-frame cost of 42.77 ms and 14.99 ms. This distributed workload enables IRIS-SLAM to balance dense semantic mapping with real-time performance, achieving an overall throughput of 13.9 FPS.

\subsection{Clustering Operator Implementation Details}
IGGT\cite{li2025iggtinstancegroundedgeometrytransformer} employs HDBSCAN\cite{mcinnes2017hdbscan} to extract instance masks $\mathcal{M}$ from the  multi-view consistent instance feature maps $\{ \mathbf{F}_t \in \mathbb{R}^{H \times W \times D} \}_{t=1}^N$. However, this approach suffers from high computational latency and an excessive GPU memory footprint, failing to meet the real-time and long-sequence processing requirements essential for SLAM systems. To address these limitations, we define a clustering operator $\beta(\mathbf{F}_t, \epsilon)$, formulated as a greedy iterative clustering algorithm designed for efficient and rapid instance mask extraction. First, the instance feature maps $\mathbf{F}_t$ are flattened and $L_2$-normalized to form the instance feature set $\mathcal{V}$. Subsequently, the algorithm randomly selects an instance feature $\mathbf{f}_s$ from $\mathcal{V}$ as the seed point, establishing an initial mask  $m_{\text{init}}$ based on the cosine similarity:
\begin{equation}
m_{\text{init}} = \{ \mathbf{f}_i \in \mathcal{V} \mid S(\mathbf{f}_i , \mathbf{f}_s) > \epsilon \},
\end{equation}
where $S(\mathbf{f}_i , \mathbf{f}_s)$ denotes the cosine similarity and $\epsilon$ is a predefined cosine similarity threshold.

To mitigate the potential initialization bias caused by random seed selection, we update the cluster center by computing the mean of all features in $m_{\text{init}}$ and applying $L_2$-normalization, resulting in the refined center $\boldsymbol{\mu}_s$:
\begin{equation}
\boldsymbol{\mu}_s = \frac{\sum_{\mathbf{f}_i \in m_{\text{init}}} \mathbf{f}_i}{\left\| \sum_{\mathbf{f}_i \in m_{\text{init}}} \mathbf{f}_i \right\|_2}.
\end{equation}
Based on this refined center $\boldsymbol{\mu}_s$, we perform a re-clustering step to determine the final instance mask $m_s$:
\begin{equation}
m_s = \{ \mathbf{f}_i \in \mathcal{V} \mid S(\mathbf{f}_i , \boldsymbol{\mu}_s) > \epsilon \}.
\end{equation}
If the number of features contained in $m_s$ exceeds the minimum threshold $\delta$, it is added to the instance masks $\mathcal{M}$; otherwise, it is discarded as background noise. In both cases, the corresponding features are removed from the instance feature set $\mathcal{V}$, and will not be reconsidered in subsequent clustering steps, resulting in a greedy peeling process. The above process repeats until $\mathcal{V}$ becomes empty.
We provide a quantitative comparison of clustering inference times between our method and IGGT in Table \ref{tab:clustering_time}.

\end{document}